\PassOptionsToPackage{table}{xcolor}
\documentclass[pdflatex,sn-mathphys-num,iicol]{sn-jnl}

\usepackage[utf8]{inputenc}
\usepackage[T1]{fontenc}
\usepackage{graphicx}
\usepackage{amsmath,amssymb,amsfonts}
\usepackage{mathrsfs}
\usepackage{textcomp}
\usepackage{xcolor}
\usepackage{xspace}
\usepackage{microtype}
\usepackage{booktabs}
\usepackage{multirow}
\usepackage{makecell}
\usepackage{tabularx}
\usepackage{adjustbox}
\usepackage{array}
\usepackage{wrapfig}
\usepackage{placeins}
\usepackage{dblfloatfix}
\usepackage{enumitem}
\usepackage{pifont}
\usepackage[normalem]{ulem}
\usepackage{caption}
\usepackage{subcaption}
\usepackage{listings}
\usepackage[most]{tcolorbox}
\usepackage{ragged2e}

\DeclareUnicodeCharacter{2019}{'}

\definecolor{eccvblue}{rgb}{0.12,0.49,0.85}
\definecolor{lightgray}{rgb}{0.8,0.8,0.8}
\definecolor{darkgreen}{rgb}{0.00,0.81,0.78}
\definecolor{gray_tab}{RGB}{220,220,220}
\definecolor{blue_tab}{RGB}{227,240,251}
\definecolor{oran_tab}{RGB}{252,242,237}
\definecolor{whit_tab}{RGB}{255,255,255}
\definecolor{green_code}{RGB}{55,126,34}

\newcommand{\xmark}{\ding{56}\xspace}

\newcommand{\method}{UniICL}
\newcommand{\dataset}{UniICL-760K}
\newcommand{\benchmark}{UniICL-Bench}

\AtBeginDocument{%
  \setlength{\abovedisplayskip}{2pt plus 1pt minus 1pt}%
  \setlength{\belowdisplayskip}{2pt plus 1pt minus 1pt}%
  \setlength{\abovedisplayshortskip}{1pt plus 1pt minus 1pt}%
  \setlength{\belowdisplayshortskip}{1pt plus 1pt minus 1pt}%
}


\newtheoremstyle{plain-tight}{6pt plus 2pt minus 1pt}{6pt plus 2pt minus 1pt}{\itshape}{}{\bfseries}{.}{.5em}{}
\theoremstyle{plain-tight}
\newtheorem{assumption}{Assumption}
\newtheorem*{proposition}{Proposition}

\makeatletter
\def\@begintheorem#1#2[#3]{%
  \deferred@thm@head{\the\thm@headfont \thm@indent
    \@ifempty{#1}{\let\thmname\@gobble}{\let\thmname\@iden}%
    \@ifempty{#2}{\let\thmnumber\@gobble}{\let\thmnumber\@iden}%
    \@ifempty{#3}{\let\thmnote\@gobble}{\let\thmnote\@iden}%
    \thm@swap\swappedhead\thmhead{#1}{#2}{#3}%
    \the\thm@headpunct
    \thmheadnl
    \hskip\thm@headsep
  }%
  \itshape
  \ignorespaces
}
\makeatother

\usepackage[capitalize]{cleveref}
\crefname{section}{Sec.}{Secs.}
\Crefname{section}{Section}{Sections}
\crefname{table}{Tab.}{Tabs.}
\Crefname{table}{Table}{Tables}
\crefname{subtable}{Tab.}{Tabs.}
\Crefname{subtable}{Table}{Tables}
\crefname{equation}{Eq.}{Eqs.}
\Crefname{equation}{Equation}{Equations}
\crefname{figure}{Fig.}{Figs.}
\Crefname{figure}{Figure}{Figures}
\crefname{subfigure}{Fig.}{Figs.}
\Crefname{subfigure}{Figure}{Figures}

\begin{document}

\title[UniICL]{UniICL: Systematizing Unified Multimodal In-Context Learning through a Capability-Oriented Taxonomy}

\author[1]{\fnm{Yicheng} \sur{Xu}}
\equalcont{These authors contributed equally to this work.}
\author*[1]{\fnm{Jiangning} \sur{Zhang}}\email{186368@zju.edu.cn}
\equalcont{These authors contributed equally to this work.}
\author[1]{\fnm{Zhucun} \sur{Xue}}
\author[2]{\fnm{Teng} \sur{Hu}}
\author[2]{\fnm{Ran} \sur{Yi}}
\author[3]{\fnm{Xiaobin} \sur{Hu}}
\author[1]{\fnm{Yong} \sur{Liu}}
\author[4]{\fnm{Dacheng} \sur{Tao}}

\affil[1]{\orgname{Zhejiang University}, \orgaddress{\country{China}}}
\affil[2]{\orgname{Shanghai Jiaotong University}, \orgaddress{\country{China}}}
\affil[3]{\orgname{National University of Singapore}, \orgaddress{\country{Singapore}}}
\affil[4]{\orgname{Nanyang Technological University}, \orgaddress{\country{Singapore}}}

\abstract{In-context learning (ICL) enables fast task adaptation from demonstrations without per-task parameter updates but remains highly sensitive to example selection and formatting. In unified multimodal models spanning understanding and generation, this sensitivity is exacerbated by cross-modal interference and varying cognitive demands. Consequently, in-context learning efficacy is often non-monotonic and highly task-dependent. To diagnose these behaviors, we introduce a six-level Capability-Oriented Taxonomy that categorizes the functional role of demonstrations from basic perception to high-order discernment. Guided by this cognitive framework, we construct {\dataset}, a large-scale corpus featuring curated 8-shot in-context learning episodes across 15 subtasks, alongside {\benchmark} for rigorous, controlled evaluation. We show that this data-driven assembly is the primary source of our gains. As a complementary, lightweight stabilizer, we additionally propose the Context-Adaptive Prototype Modulator, a plug-and-play module that further improves few-shot stability. Evaluations on {\benchmark} show that our approach yields highly competitive unified results, outperforming larger-parameter multimodal large language model baselines on most understanding in-context learning tasks. Data and code are available at \url{https://github.com/xuyicheng-zju/UniICL}.}


\maketitle

\section{Introduction}
\label{sec:intro}
\begin{figure*}[t]
    \centering
    \includegraphics[width=\textwidth]{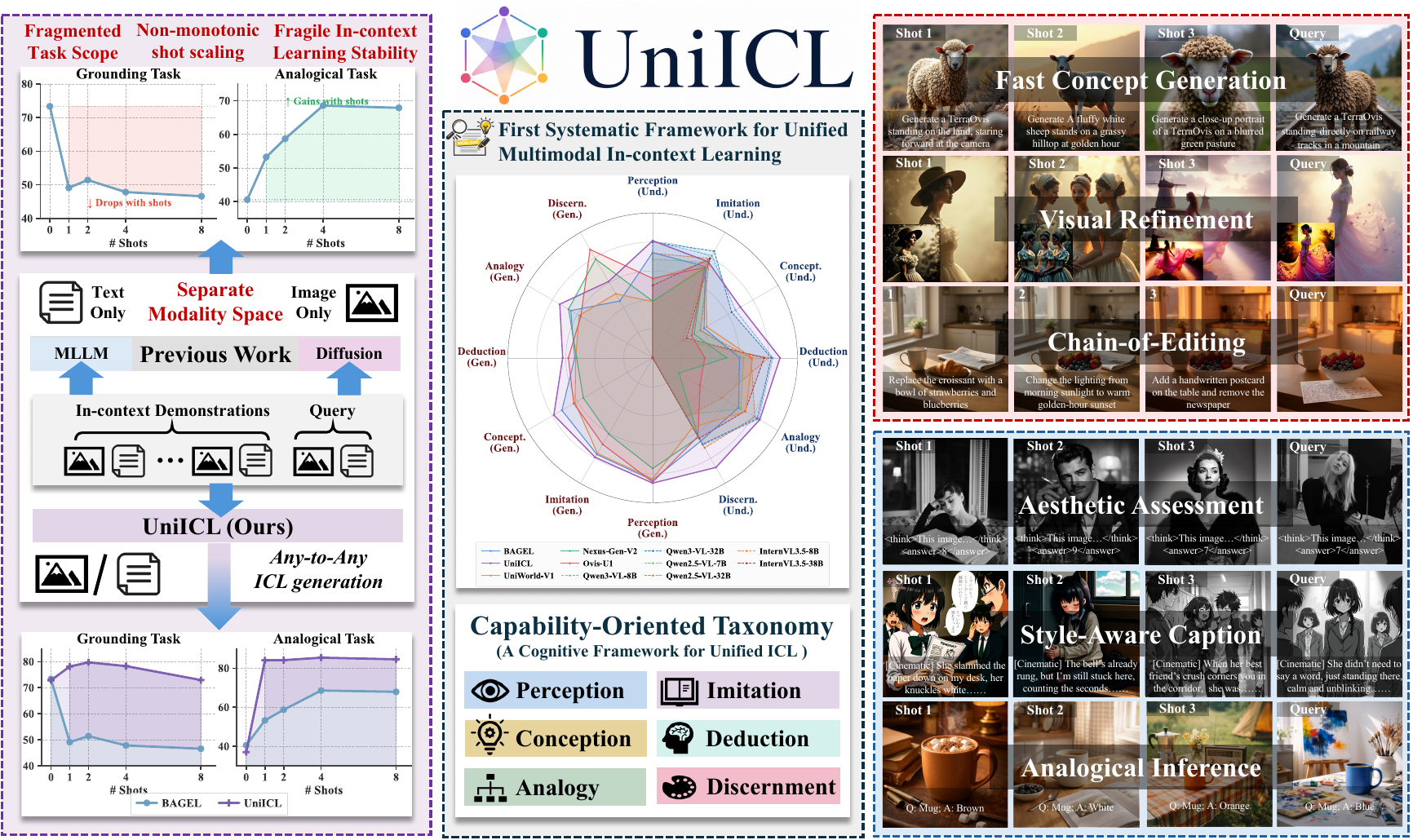}
    \caption{
    \textbf{\textit{Left:}} Previous fragmented paradigms isolate modalities and tasks, often suffering from \emph{non-monotonic shot scaling}. Our {\method} mitigates this issue to achieve consistent gains.
    \textbf{\textit{Middle:}} Our six-level Capability-Oriented Taxonomy and a radar chart across understanding and generation tasks. 
    \textbf{\textit{Right:}} ICL examples from {\dataset}.
    }
    \label{fig:teasor}
\end{figure*}
In-context learning (ICL) enables systems to perform novel tasks via few-shot demonstrations without per-task updates~\cite{brown2020language}. This paradigm is increasingly applied to multimodal systems~\cite{alayrac2022flamingo,li2023blip,li2025otter,li2025m2iv} and image synthesis~\cite{li2025visualcloze,koh2023generating,dong2023dreamllm}. As architectures evolve toward unified models jointly supporting understanding and generation, integrating robust ICL across disparate tasks within a shared interface becomes exceptionally difficult. Dense interleaving of text and visual tokens introduces cross-modal interference, making unified ICL highly sensitive to demonstration selection and modality balance~\cite{qin2024factors,lu2022fantastically,chen2025can,zhao2021calibrate}.

Existing work often studies multimodal ICL through task- or modality-specific lenses (\cref{fig:teasor}, left). Most benchmarks focus on zero-shot evaluation, while existing ICL datasets remain modality asymmetric, favor visual question answering (VQA) over generative tasks~\cite{li2023mimicit}, and lack a systematic cognitive structure~\cite{zhao2023mmicl}. This mixes low-level perceptual anchors with abstract analogical scaffolds~\cite{alayrac2022flamingo,li2023blip,li2025visualcloze}, obscuring how ICL scales under different cognitive loads and leaving \emph{non-monotonic shot scaling} (\cref{fig:teasor}, top-left) largely unexplored. In practice, demonstrations can degrade perception-dominant tasks through visual distraction while reinforcing structural patterns for complex inductive tasks.

Diagnosing these failure modes requires a capability-oriented perspective. As our fundamental contribution, we introduce a six-level Capability-Oriented Taxonomy (\cref{fig:teasor}, center) inspired by neurocognitive development~\cite{craik1972levels,felleman1991distributed,halford1998processing}. It structures ICL tasks by the functional role of demonstrations: perception, imitation, conception, deduction, analogy, and discernment. Guided by this taxonomy, we construct {\dataset}, a large-scale unified multimodal ICL dataset with over 766,000 curated episodes across 15 subtasks (partially illustrated in \cref{fig:teasor}, right). We further derive {\benchmark}, the first cognitively structured testbed for evaluating multi-dimensional ICL capabilities, stability, and controlled 0--8-shot behavior.

While our structured dataset establishes the foundation for unified multimodal in-context learning, standard self-attention remains susceptible to cross-modal noise in dense contexts. As a complementary, lightweight stabilizer, we further propose the Context-Adaptive Prototype Modulator (CAPM), a plug-and-play module that disentangles demonstration encoding and adaptively routes context, providing additional few-shot stability on top of the data-driven gains. In summary, our core contributions are threefold:
\begin{itemize}[label=\textbullet, wide=0pt, leftmargin=*, labelsep=0.5em]
    \item We introduce a six-level Capability-Oriented Taxonomy for multimodal ICL evaluation, exposing non-monotonic scaling behaviors in unified models.
    \item We present {\dataset} and {\benchmark}, a cognitively structured training corpus whose demonstration-assembly pipeline is the main source of our gains, together with a rigorous evaluation suite for unified understanding and generative ICL.
    \item We propose CAPM, a lightweight auxiliary module that further stabilizes few-shot adaptation. Combined with our data, the full approach matches or surpasses larger-parameter multimodal large language models (MLLMs) on most understanding tasks.
\end{itemize}

\section{Related Work}

\subsection{Multimodal In-Context Learning}
In-context learning emerged as a foundational capability in large language models such as GPT-3~\cite{brown2020language}, enabling adaptation from demonstrations through mechanisms including implicit gradient descent~\cite{von2023transformers}, Bayesian inference~\cite{xie2021explanation}, and induction heads~\cite{olsson2022context}. Extending this capability to the visual domain, systems such as Flamingo~\cite{alayrac2022flamingo}, BLIP-2~\cite{li2023blip}, and IDEFICS~\cite{laurenccon2024matters} use interleaved image-text demonstrations as grounding cues for perception-centric tasks like visual question answering and image captioning. Parallel research in visual synthesis, including Visualcloze~\cite{li2025visualcloze} and CoDi-2~\cite{tang2024codi}, frames in-context conditioning as spatial editing and style transfer. These lines of research approach in-context learning through task-specific perspectives, optimizing for either text-output perception or pixel-level manipulation in isolation. To address this fragmentation, our Capability-Oriented Taxonomy organizes context-dependent demands across six levels within a unified evaluation framework that covers both understanding and generation.

\subsection{Datasets and Benchmarks for Multimodal ICL}
The development of multimodal datasets initially focused on training and evaluating zero-shot capabilities. While NLP instruction-tuning collections~\cite{wang2022super,longpre2023flan} established standardized evaluation protocols, multimodal benchmarks such as MMBench~\cite{liu2024mmbench}, MMMU~\cite{yue2024mmmu}, and SEED-Bench~\cite{li2023seed} extended this evaluation to vision-language reasoning. Existing datasets for multimodal in-context learning generally separate understanding from generation. They are often modality-asymmetric, favoring visual question answering over generative tasks, and lack a unified cognitive structure by blending low-level perception with complex deduction. This structural deficiency obscures the specific scaling behaviors of in-context learning under varying cognitive loads. To address these gaps, we propose UniICL-760K as the first large-scale dataset for unified understanding and generation guided by a systematic taxonomy. From this collection, we derive UniICL-Bench, a cognitively structured benchmark for unified multimodal ICL and few-shot stability.

\subsection{Unified Multimodal Foundation Models}
Unified multimodal models integrate understanding and generation within a single autoregressive backbone. Systems such as GILL~\cite{koh2023generating}, DreamLLM~\cite{dong2023dreamllm}, Chameleon~\cite{team2024chameleon}, and the Emu series~\cite{sun2023emu,sun2024generative,wang2024emu3} achieve interleaved multimodal input and output by aligning visual tokens with discrete text vocabularies. Recent models, including BAGEL~\cite{deng2025emerging}, UniWorld-V1~\cite{lin2025uniworld}, Nexus-Gen-V2~\cite{zhang2025nexus}, and Ovis-U1~\cite{wang2025ovis}, advance these capabilities while retaining similar architectural foundations. The paradigm of unified multimodal in-context learning aims to achieve broad generalization across diverse tasks using a single model. Unifying these processes introduces specific challenges compared to fragmented approaches. Models face inherent optimization tensions, such as modality competition and the alignment tax caused by heterogeneous token spaces, frequently resulting in unstable few-shot in-context learning behaviors. We address these issues primarily through a cognitively structured data-construction pipeline that supplies well-assembled demonstrations, and complement it with a lightweight modulation module (CAPM) that further stabilizes few-shot adaptation.

\section{Methodology: Dataset, Benchmark, and Model}
\label{sec:methodology}

\subsection{Formulation of Unified Multimodal In-Context Learning}
\label{sec:task_formulation}

Unified multimodal in-context learning adapts foundation models to diverse understanding and generation tasks without parameter updates. Beyond architectural unification, it consolidates multimodal demands into a single cohesive prompting framework. We formalize this as a universal conditional prediction problem, where an ICL episode comprises a $k$-shot context $\mathcal{D}=\{d_i\}_{i=1}^{k}$ and a target query $(x^\star, q^\star)$. Each demonstration is a triplet $d_i=(x_i, q_i, y_i)$ with visual input $x_i$, textual instruction $q_i$, and ground-truth output $y_i$, where $x_i=\varnothing$ for text-only cases. Operating over a shared text-image space $\mathcal{Y}$, the model predicts $y^\star \in \mathcal{Y}$ conditioned on the context:
$$y^\star \sim p_\theta(\cdot \mid \mathcal{D}, x^\star, q^\star).$$
This interface naturally supports mixed-modality episodes, determining the output modality via $q^\star$. However, unified adaptation introduces significant challenges: \textit{highly variable task formats and dense text-image interleaving frequently trigger severe cross-modal interference.} Consequently, models exhibit high sensitivity to demonstration selection, causing unstable, non-monotonic scaling. Resolving this requires a capability-oriented framework to diagnose failure modes alongside robust algorithmic interventions for stabilization.

\subsection{Curating {\dataset} Dataset}
\label{sec:dataset}
\begin{figure*}[!t]
    \centering
    \includegraphics[width=\textwidth]{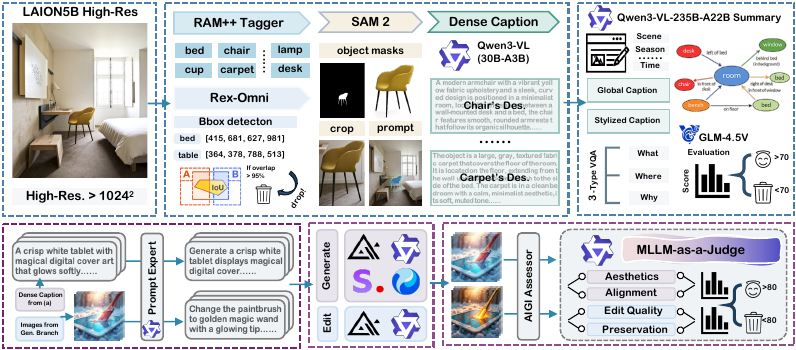}
    \caption{\textbf{{\dataset} data-asset curation pipeline.} The annotation branch builds validated dense annotations from LAION-HR, while the synthesis branch expands high-quality generation and editing assets through expert models and quality filtering.}
    \label{fig:pipeline}
\end{figure*}

We introduce \textbf{{\dataset}}, a large-scale dataset specifically designed for unified multimodal in-context learning across visual understanding and generation. It contains 766,868 carefully constructed ICL episodes, each paired with a curated 8-shot demonstration context. These episodes are assembled from two source branches: 202,750 validated scene-centric annotations and 353,826 quality-controlled synthetic assets, which are reused across multiple tasks to form the final episode set. Rather than fragmenting tasks by isolated application goals, {\dataset} organizes understanding and generation within a six-level Capability-Oriented Taxonomy, instantiating 15 corresponding subtasks. Due to the high cost of constructing expert-level editing trajectories, the \emph{Chain-of-Editing} subtask is excluded from the training corpus, retained solely in our benchmark to evaluate generative generalization. Overall, {\dataset} serves as a scalable training resource for unified multimodal ICL, while the independently curated {\benchmark} enables systematic evaluation.
\subsubsection{Taxonomy-Guided ICL Task Instantiation}
Demonstrations in an ICL episode serve qualitatively different roles depending on the underlying task. To systematically categorize these functional roles, we introduce a \textbf{\textit{six-level Capability-Oriented Taxonomy}} inspired by the neurocognitive progression from shallow perceptual analysis to deep semantic reasoning~\cite{craik1972levels,felleman1991distributed,halford1998processing}. The taxonomy is defined by \emph{what demonstrations contribute to the solution process}, not by surface task format, and the 15 subtasks below are designed to cover these functional roles. We later show that same-taxonomy tasks share shot-scaling behavior while cross-taxonomy tasks do not (\cref{fig:taxonomy_trend_corr} in Sec.~\ref{sec:experiments}), supporting the distinction as empirically meaningful.

\begin{enumerate}[label=\textbf{\arabic*.}, wide=0pt, leftmargin=*, labelsep=0.5em, itemsep=0.2em]
    \item \textbf{Perception.} At this foundational level, context serves as an explicit perceptual anchor. The model must selectively attend to targeted visual evidence defined by demonstrations, resisting irrelevant distractors or hallucinated priors. We evaluate this fine-grained attentional allocation and spatial localization through \emph{Visual Grounding}, \emph{Attribute Recognition}, and \emph{Image Manipulation}.
    \item \textbf{Imitation.} Moving beyond passive perception, this level evaluates active observational learning. Demonstrations act as instructors, requiring the model to internalize and reproduce specific structural, stylistic, or logical schemas. We instantiate this capability through \emph{Style-Aware Caption}, \emph{Scene Reasoning}, and \emph{Instructional Generation}. Although Imitation and Perception both rely on visual evidence, they differ in how context is used: Perception treats it as a passive anchor to select against, while Imitation treats it as an active schema to reproduce.
    \item \textbf{Conception.} This level assesses the core fast-mapping capability of robust ICL: rapidly binding novel linguistic symbols to unseen visual concepts. Context introduces counterfactual or out-of-distribution concepts, demanding that the model temporarily adopt definitions established within the episode. We map this cognitive demand to \emph{Fast Concept Mapping} and \emph{Fast Concept Generation}.
    \item \textbf{Deduction.} This level demands extracting causal or temporal sequences across multiple demonstrations. Rather than isolated mappings, context establishes progressive multi-step coherence rules that the model must deduce to resolve the target query. We measure this capacity through \emph{World-Aware Planning} and \emph{Chain-of-Editing}.
    \item \textbf{Analogy.} This category evaluates abstract generalization, requiring the transfer of hidden transformation rules across diverse surface forms~\cite{gentner1983structure}. Given uninstructed demonstrations sharing an underlying intent, the model must autonomously extract the implicit rule. We assess this through \emph{Analogical Inference} and \emph{Analogical Editing}, applying derived rules to novel visual domains.
    \item \textbf{Discernment.} At our highest cognitive level, this category evaluates value judgment. Context conveys human-aligned criteria for beauty, authenticity and creation rather than deterministic rules. We assess this capability through \emph{Aesthetic Assessment}, \emph{Forgery Detection}, and \emph{Visual Refinement}. Demonstrations therefore include explicit criteria and short rationales rather than only final judgments.
\end{enumerate}

\begin{figure*}[!t]
    \centering
    \begin{subfigure}[t]{0.49\textwidth}
        \centering
        \includegraphics[width=\textwidth]{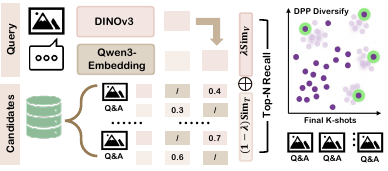}
        \caption{\textbf{Feature-Based assembly.}}
        \label{fig:feature_retrieval}
    \end{subfigure}
    \hfill
    \begin{subfigure}[t]{0.49\textwidth}
        \centering
        \raisebox{-1ex}[\height][\depth]{\includegraphics[width=\textwidth]{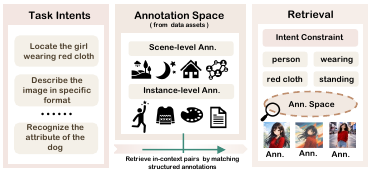}}
        \caption{\textbf{Intent-Based assembly.}}
        \label{fig:intent_assembly}
    \end{subfigure}
    \caption{\textbf{Task-aligned demonstration assembly.} Feature-Based assembly retrieves relevant and diverse examples in multimodal embedding space. Intent-Based assembly retrieves examples by matching structured task constraints against annotations.}
    \label{fig:assembly}
\end{figure*}

\subsubsection{{\method} Curation Pipeline}
\label{sec:curation_pipeline}

Constructing multimodal ICL datasets is inherently costly, requiring rigorous query-demo pairing. Prior works~\cite{alayrac2022flamingo,awadalla2023openflamingo} often rely on basic visual similarity for filtering, which limits task coverage: many ICL tasks require structural alignment or abstract rule extraction beyond shared global visual semantics. To curate a large-scale dataset spanning diverse task demands, we first construct high-quality data assets (\cref{fig:pipeline}) and then assemble task-aligned demonstrations (\cref{fig:assembly}).

\begin{figure*}[!t]
    \centering
    \begin{subfigure}[t]{0.49\textwidth}
        \centering
        \includegraphics[width=\textwidth]{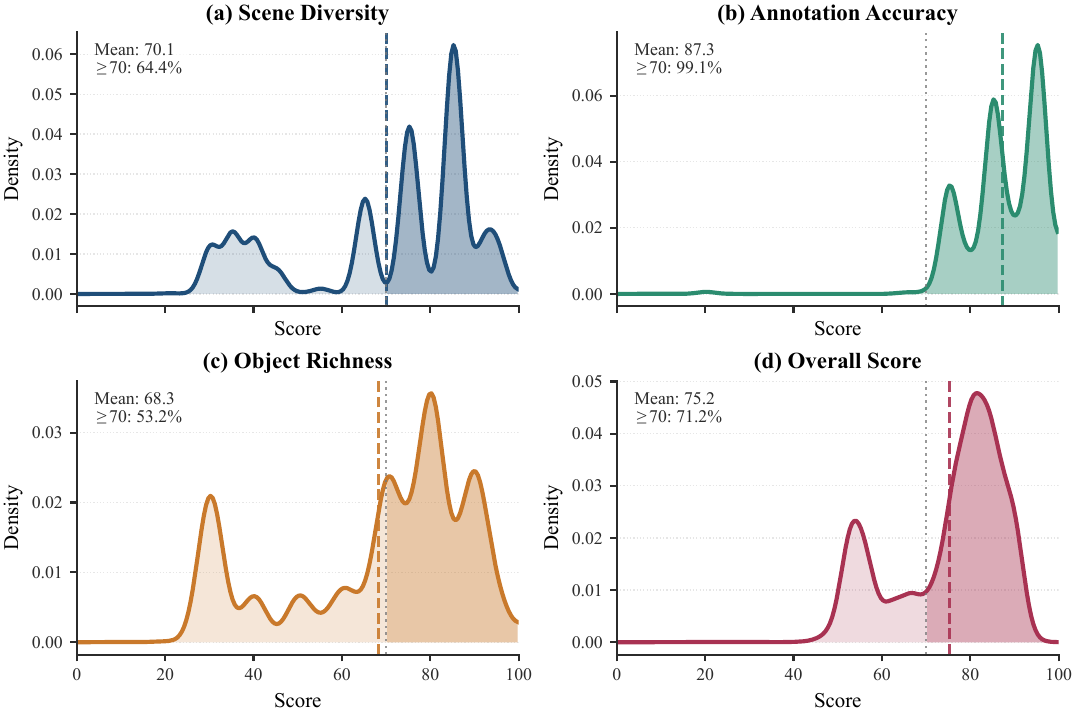}
        \caption{\textbf{Annotation-side score distributions.}}
        \label{fig:quality_dist}
    \end{subfigure}
    \hfill
    \begin{subfigure}[t]{0.49\textwidth}
        \centering
        \includegraphics[width=\textwidth]{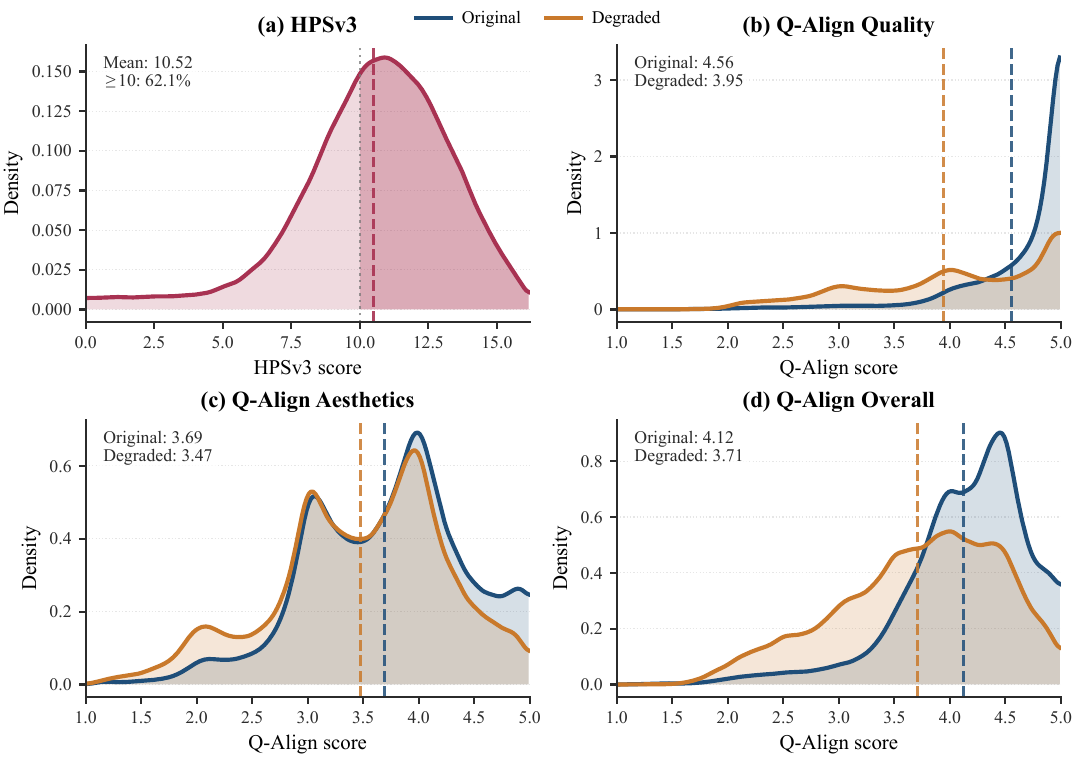}
        \caption{\textbf{Synthetic-side score distributions.}}
        \label{fig:t2i_dist}
    \end{subfigure}
    \caption{\textbf{Score distributions for data-asset filtering.} Annotation-side filtering is bottlenecked by scene diversity and object richness. Synthetic-side filtering uses HPSv3 for instruction quality and Q-Align for refinement quality gaps.}
    \label{fig:score_dist}
\end{figure*}

\begin{figure*}[!t]
    \centering
    \includegraphics[width=\textwidth]{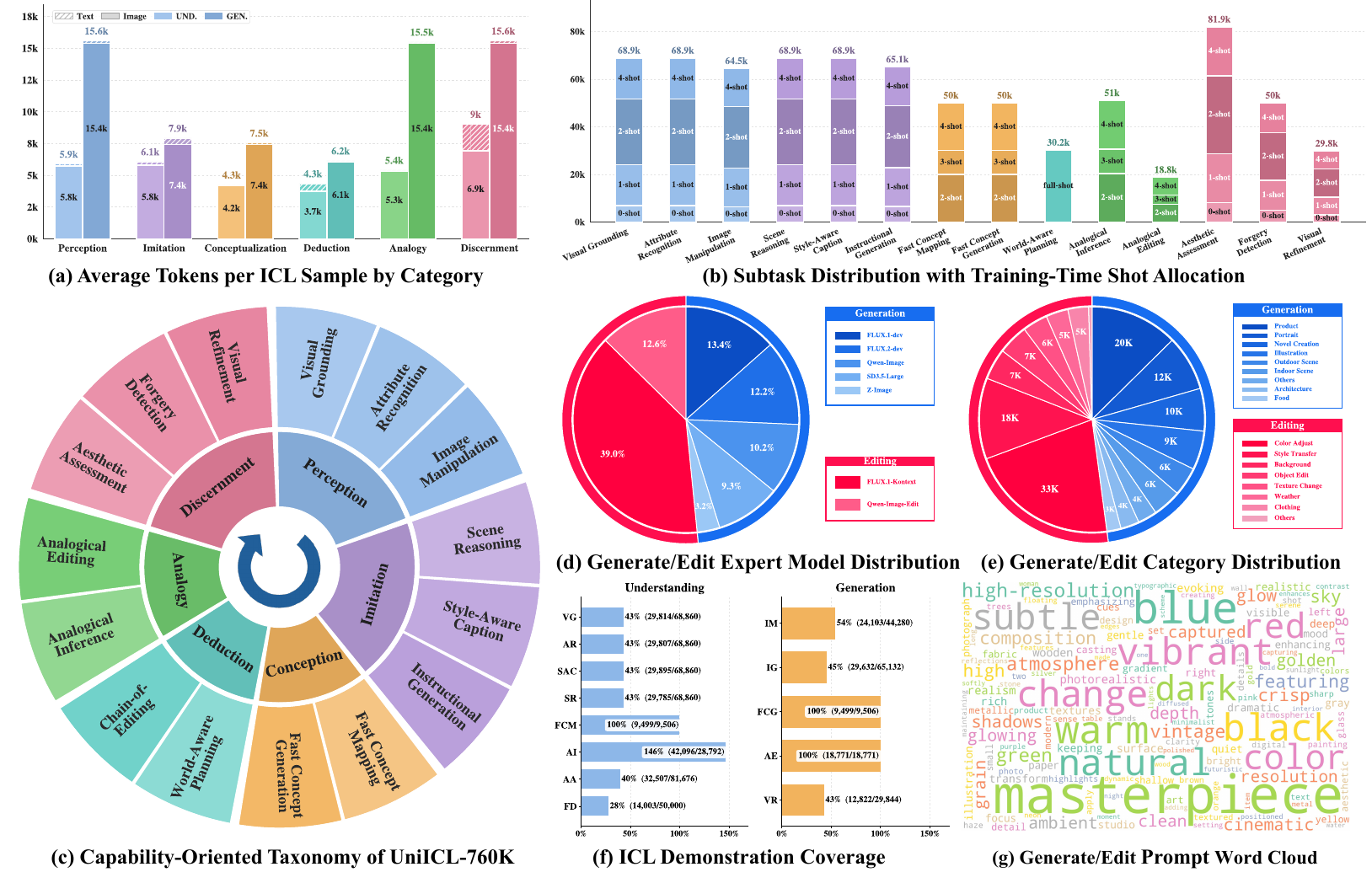}
    \caption{\textbf{Statistical distributions of our {\dataset} from multiple perspectives.}}
    \label{fig:state}
\end{figure*}

\noindent\textbf{(1) Automatic Data Asset Construction.} 
We build a highly structured semantic repository via two complementary approaches. 
\textbf{\textit{(a)} Cascaded Dense Annotation.} To extract fine-grained semantics, we curate a high-resolution LAION-5B~\cite{schuhmann2022laion} subset through a six-stage cascaded pipeline: \textbf{(S1)} open-vocabulary tagging with RAM++~\cite{zhang2024recognize}, \textbf{(S2)} detection-consistency filtering with Rex-Omni~\cite{jiang2025detectpointprediction}, which keeps a sample only when box- and point-based category counts agree, \textbf{(S3)} instance segmentation with SAM~2~\cite{ravi2024sam} plus IoU-based deduplication, yielding high-fidelity masks and dual-view representations, \textbf{(S4)} instance-level dense captioning and attribute extraction with Qwen3-VL-30B-A3B-Instruct~\cite{bai2025qwen3}, \textbf{(S5)} corrective scene-level aggregation with Qwen3-VL-235B-A22B-Instruct~\cite{bai2025qwen3} for global attributes, stylized captions, scene graphs, and VQA pairs, and \textbf{(S6)} hallucination validation with GLM-4.5V~\cite{hong2025glm}. Stages S2 and S6 are the main filters, reducing the 750k raw-image source to 283,838 validated scene annotations, from which 202,750 high-quality samples survive the final quality threshold. The stage-wise retention in \cref{tab:data_asset_stats}(a) shows that scene diversity and object richness are the binding constraints, while annotation accuracy is already saturated.
\textbf{\textit{(b)} Cascaded Generative Synthesis.} To synthesize high-quality images for generative ICL tasks, we expand our dataset using advanced generative and editing expert models (\cref{fig:state}-d). We employ Qwen3-VL-8B-Instruct~\cite{bai2025qwen3} as a unified prompt expert. For text-to-image synthesis, it translates dense scene captions from the annotation branch into detailed generation prompts, while generating context-aware instructions for image-to-image editing pairs. All outputs are rigorously filtered by Q-Align~\cite{wu2023q}, HPSv3~\cite{ma2025hpsv3}, and a multimodal large language model judge (MLLM-Judge)~\cite{zheng2023judging} to retain high-fidelity data. As detailed in \cref{tab:data_asset_stats}(b) and the HPSv3/Q-Align score distributions in \cref{fig:t2i_dist}, the instruction pool is gated by HPSv3 ($>10$) and the refinement pool is defined by a Q-Align quality gap ($>0.5$) between original and degraded images. The four pools reuse upstream synthetic sources, with the instruction pool seeding both edit and refinement, and are not additive. The refinement pool counts degraded-to-clean pairs, and the de-duplicated asset pool, comprising 99,455 instruction-following images, 81,202 edited images, 97,683 degraded-to-clean refinement pairs, and 11,050 concept-oriented synthetic images. The specialized branches built from AVA~\cite{murray2012ava}, AIGI-Holmes~\cite{zhou2025aigi}, and World-Aware Planning~\cite{shi2025worldawareplanningnarrativesenhance} require task-specific processing, which is detailed in the Supplementary Material.

\begin{table}[!b]
    \caption{\textbf{Data-asset construction statistics.}}
    \label{tab:data_asset_stats}
    \centering
    \footnotesize
    \renewcommand{\arraystretch}{1.15}
    \begin{tabular*}{\columnwidth}{@{}c@{}}
    \begin{minipage}{\columnwidth}
    \centering
    \textbf{(a) Annotation-pipeline retention statistics.}\par
    \vspace{0.25em}

    \setlength{\tabcolsep}{6pt}
    \begin{tabular*}{\columnwidth}{@{\extracolsep{\fill}}lcc@{}}
        \toprule
        \textbf{Stage} & \textbf{Retained} & \textbf{Rate} \\
        \midrule
        Raw input                               & 750,000 & 100.0\% \\
        After detection filtering (S2)          & 521,921 & 69.6\% \\
        After dense captioning (S4)             & 521,656 & 69.6\% \\
        After validity check (S6)               & 283,838 & 37.8\% \\
        \midrule
        Scene diversity $>70$                   & 183,567 & 24.5\% \\
        Annotation accuracy $>70$               & 282,259 & 37.6\% \\
        Object richness $>70$                   & 151,515 & 20.2\% \\
        Overall score $>70$                     & 202,750 & 27.0\% \\
        \bottomrule
    \end{tabular*}

    \vspace{0.65em}
    \textbf{(b) Synthetic-asset construction statistics.}\par
    \vspace{0.25em}

    \setlength{\tabcolsep}{4.5pt}
    \begin{tabular*}{\columnwidth}{@{\extracolsep{\fill}}lccc@{}}
        \toprule
        \textbf{Asset pool} & \textbf{Source} & \textbf{Generated} & \textbf{Filter} \\
        \midrule
        Instruction pool & 163,891 & 99,455  & HPSv3 $>10$ \\
        Edit pool        & 99,455  & 81,202  & MLLM-Judge \\
        Refinement pool  & 99,455  & 97,683  & Q-Align gap $>0.5$ \\
        Concept pool     & 10,000  & 11,050  & MLLM-Judge \\
        \bottomrule
    \end{tabular*}
    \end{minipage}
    \end{tabular*}
\end{table}

\noindent\textbf{(2) Task-Aligned Demonstration Retrieval.}
Effective $k$-shot episodes require retrieving contextual demonstrations aligned with the available multimodal query intent $q = (x_q, \tau_q)$, with $\tau_q$ as the query-side instruction or intent descriptor rather than the target output. We adopt two complementary \emph{assembly} paradigms accommodating varying task abstractions: \textbf{Feature-Based assembly}, which uses multimodal similarity retrieval for Perception, Imitation, and Discernment tasks, and \textbf{Intent-Based assembly}, which uses logical metadata matching for Conception, Deduction, and Analogy tasks. The task-level mapping is listed in \cref{tab:benchmark_scale}.
\textbf{\textit{(c)} Feature-Based Assembly.} For tasks relying on semantic alignment, we compute a fused cross-modal similarity between the query $q$ and candidate demonstrations $d = (x_d, \tau_d)$, where $\tau_d$ is the candidate-side text descriptor used for retrieval, as illustrated in \cref{fig:feature_retrieval}. Let DINOv3~\cite{simeoni2025dinov3} be $E_v$ and Qwen3-Embedding~\cite{zhang2025qwen3} be $E_t$. The fusion score $\mathcal{S}(q, d)$ is defined as:
\begin{equation}
\begin{aligned}
    \mathcal{S}(q, d) =
    &\lambda \frac{E_v(x_q) \cdot E_v(x_d)}
    {\|E_v(x_q)\| \|E_v(x_d)\|} \\
    &+ (1 - \lambda) \frac{E_t(\tau_q) \cdot E_t(\tau_d)}
    {\|E_t(\tau_q)\| \|E_t(\tau_d)\|},
\end{aligned}
\end{equation}
where $\lambda{=}0.5$ balances modality preference. To ensure relevance and diversity, we frame context selection as a Determinantal Point Process (DPP)~\cite{taskar2013determinantal}. For top-$N$ candidates, we construct a positive semi-definite L-ensemble kernel $\mathbf{L}_{ij} = w_i w_j \cdot \phi_i^\top \phi_j$ with quality weights $w_i = \exp(\beta \cdot s_i),\ \beta{=}8$, where $\phi_i$ is the $\ell_2$-normalized DINOv3 visual feature of shot $d_i$ and $s_i$ is its multimodal relevance score. This factorizes as $\mathbf{L} = \mathbf{B}\mathbf{B}^\top$ with $\mathbf{B}_i = w_i\phi_i$. We use a greedy Cholesky approximation~\cite{taskar2013determinantal} to maximize $\det(\mathbf{L}_Y)$, selecting at each step the candidate with maximum residual norm after projecting out already-selected directions.
\textbf{\textit{(d)} Intent-Based Assembly.} Certain implicit inductive tasks resist continuous visual representation, as illustrated in \cref{fig:intent_assembly}. For instance, global semantics poorly capture tasks requiring precise coordinates, such as a woman wearing red. Here, we formalize the query intent as a Boolean conceptual rule $R$. In this example, the logical conjunction is $R = (\text{category} = \text{woman}) \land (\text{color} = \text{red}) \land (\text{coord} \in \Omega_{\text{target}})$, where $\Omega_{\text{target}}$ defines the target region. Bypassing the latent space, we evaluate this logic directly against the fine-grained metadata $\mathcal{M}_j$ associated with each candidate demonstration $d_j$. The context subset is retrieved via exact logical satisfaction:
\begin{equation}
    \mathcal{D}_{\text{target}} = \big\{ d_j \in \mathcal{D} \mid \mathcal{M}_j \models R \big\},
\end{equation}
where $\models$ is metadata satisfaction of the Boolean rule $R$, enforcing task-specific context alignment.

\begin{table*}[!t]
\centering
\caption{\textbf{Comparison with existing multimodal benchmarks.}}
\label{tab:benchmark_comparison}
\renewcommand{\arraystretch}{1.0}
\setlength\tabcolsep{6pt}
\resizebox{\textwidth}{!}{%
\begin{tabular}{@{}c c c cc ccc ccc@{}}
\toprule
\multirow{2}{*}{\textbf{Benchmark}}
  & \multirow{2}{*}{\textbf{Scale}}
  & \multirow{2}{*}{\textbf{Year}}
  & \multicolumn{2}{c}{\textbf{Task Coverage}}
  & \multicolumn{3}{c}{\textbf{Evaluation Metrics}}
  & \multicolumn{3}{c}{\textbf{ICL-Centric Eval.}} \\
\cmidrule(lr){4-5}\cmidrule(lr){6-8}\cmidrule(l){9-11}
  & & \rule[-0.55ex]{0pt}{2.6ex}
  & \makecell[c]{\textbf{Und.}} & \makecell[c]{\textbf{Gen.}}
  & \textbf{Trad.} & \textbf{Expert} & \textbf{MLLM}
  & \textbf{0--8 Shot} & \textbf{Taxonomy} & \textbf{Stability} \\
\midrule
MMBench~\cite{liu2024mmbench}
  & 3,217  & 2023
  & \checkmark &
  & \checkmark &            & \checkmark
  &            &            &            \\
MMMU~\cite{yue2024mmmu}
  & 11,550 & 2024
  & \checkmark &
  & \checkmark &            &
  &            &            &            \\
SEED-Bench~\cite{li2023seed}
  & 19,242 & 2023
  & \checkmark & 
  & \checkmark & \checkmark &
  &            &            &            \\
GenEval~\cite{ghosh2023geneval}
  & 553      & 2023
  &            & \checkmark
  &  & \checkmark           &
  &            &            &            \\
VL-ICL-Bench~\cite{zong2024vl}
  & 1,760  & 2024
  & \checkmark & \checkmark
  & \checkmark &            & \checkmark
  & \checkmark &            &            \\
\midrule
\rowcolor{eccvblue!10}\benchmark~(Ours)
  & 1,250 & 2026
  & \checkmark & \checkmark
  & \checkmark & \checkmark & \checkmark
  & \checkmark & \checkmark & \checkmark \\
\bottomrule
\end{tabular}%
}
\end{table*}

The branch-level quality statistics are summarized in \cref{fig:state}, with the score distributions of the two curation branches shown in \cref{fig:quality_dist,fig:t2i_dist}. For annotation-side filtering, the final threshold keeps 202,750 samples. Annotation accuracy is already concentrated at a high range, with mean $87.3$ and $99.1\%$ above $70$, while scene diversity, with mean $70.1$ and $64.4\%$ above $70$, and especially object richness, with mean $68.3$ and $53.2\%$ above $70$, act as the binding bottlenecks. On the synthetic side, HPSv3 gates the instruction pool and Q-Align separates original from degraded images, retaining 99,455 high-fidelity instruction-following images, while subsequent filtering yields 81,202 edited images and 97,683 refinement pairs with clear quality gaps for supervised refinement.

\begin{table*}[!t]
\centering
\caption{\textbf{Composition of {\dataset} and {\benchmark} across the six cognitive levels.} Image Source reports the task-level image origin, and Assembly reports the demonstration-assembly paradigm. Feat. means Feature-Based and Intent means Intent-Based. Train Pool reports the deduplicated and benchmark-isolated training-pool size per task. Bench Episodes reports the evaluation count. Chain-of-Editing is benchmark-only and excluded from training.}
\label{tab:benchmark_scale}
\small
\setlength{\tabcolsep}{5pt}
\resizebox{\textwidth}{!}{
\begin{tabular}{@{}lccccccc@{}}
\toprule
\textbf{Taxonomy} & \textbf{Sub-task} & \textbf{Image Source} & \textbf{Modality} & \textbf{Assembly} & \textbf{Evaluation Metrics} & \textbf{Train Pool} & \textbf{Bench Episodes} \\
\midrule
\multirow{3}{*}{1. Perception}
 & Visual Grounding       & LAION-HR~\cite{schuhmann2022laion} & Und. & Feat. & mIoU                       & 66,347 & 100 \\
 & Attribute Recognition  & LAION-HR & Und. & Feat. & Accuracy                   & 64,338 & 100 \\
 & Image Manipulation     & Synthesis & Gen. & Feat. & MLLM-Judge                 & 59,415 & 50 \\
\midrule
\multirow{3}{*}{2. Imitation}
 & Style-Aware Caption    & LAION-HR & Und. & Feat. & MLLM-Judge / BERTScore     & 67,225 & 100 \\
 & Scene Reasoning        & LAION-HR & Und. & Feat. & MLLM-Judge / BERTScore     & 66,074 & 100 \\
 & Instructional Generation & Synthesis & Gen. & Feat. & HPSv3                     & 60,990 & 50 \\
\midrule
\multirow{2}{*}{3. Conception}
 & Fast Concept Mapping   & Synthesis & Und. & Intent & Accuracy                   & 50,000 & 100 \\
 & Fast Concept Generation & Synthesis & Gen. & Intent & MLLM-Judge                 & 50,000 & 100 \\
\midrule
\multirow{2}{*}{4. Deduction}
 & World-Aware Planning   & World-Aware Planning~\cite{shi2025worldawareplanningnarrativesenhance} & Und. & Intent & Accuracy                   & 29,824 & 100 \\
 & Chain-of-Editing       & Synthesis & Gen. & Intent & MLLM-Judge                 & --     & 50 \\
\midrule
\multirow{2}{*}{5. Analogy}
 & Analogical Inference   & LAION-HR & Und. & Intent & MLLM-Judge                 & 51,028 & 100 \\
 & Analogical Editing     & Synthesis & Gen. & Intent & MLLM-Judge / DINOv3        & 18,710 & 50 \\
\midrule
\multirow{3}{*}{6. Discernment}
 & Aesthetic Assessment   & AVA~\cite{murray2012ava} & Und. & Feat. & SRCC / PLCC                & 80,481 & 100 \\
 & Forgery Detection      & AIGI-Holmes~\cite{zhou2025aigi} & Und. & Feat. & Accuracy                   & 40,661 & 100 \\
 & Visual Refinement      & Synthesis & Gen. & Feat. & Q-Align Eff.               & 27,996 & 50 \\
\bottomrule
\end{tabular}
}
\end{table*}

\subsubsection{UniICL-Bench Construction and Evaluation}
For systematic evaluation, we construct \textbf{\benchmark}, a rigorously vetted testbed comprising 1,250 episodes distributed across all six cognitive dimensions. We keep the benchmark compact so repeated evaluation remains reproducible and accessible to the community. The same episodes still expand to 5,650 ICL instances under the shot sweep \(k\in\{0,1,2,4,8\}\), and one full model pass costs about 160 H20 GPU-hours. As outlined in \cref{tab:benchmark_comparison}, unlike existing benchmarks skewed heavily toward VQA~\cite{liu2024mmbench,yue2024mmmu} or isolated generation~\cite{ghosh2023geneval}, \benchmark{} provides unified cross-modality assessment within a shared cognitive framework. Its role is intentionally different from that of the training corpus. {\dataset} supplies large-scale supervision, whereas {\benchmark} is kept compact, balanced, and diagnostic.
The benchmark is organized around the same Capability-Oriented Taxonomy as the training set, allowing us to compare understanding and generation under a single capability-oriented view rather than as disconnected task families. As summarized in \cref{tab:benchmark_scale}, the suite spans 15 subtasks, covers both discriminative and generative settings, and preserves free-form response formats to maintain ICL contextual alignment. Multiple-choice reformulation can degenerate ICL into option matching: the model may exploit answer priors or textual cues in the choices rather than infer the transformation demonstrated by the context, making shot scaling brittle for concept formation, analogical transfer, and visual refinement. Deterministic outputs are parsed and evaluated with objective metrics such as accuracy, mIoU, SRCC, and PLCC. For semantically open-ended tasks, we combine specialized neural scorers with the MLLM-Judge protocol, since single overlap-based metrics are often unreliable for assessing concept transfer, scene reasoning, or instruction-following generation. In particular, we use HPSv3~\cite{ma2025hpsv3} for instruction-following generation, Q-Align~\cite{wu2023q} for quality-aware refinement, and MLLM-Judge where semantic equivalence matters more than exact strings. The full task-to-metric mapping is listed in \cref{tab:benchmark_scale}.

\noindent\textbf{Stability evaluation.}
Most existing benchmarks in \cref{tab:benchmark_comparison} report clean-context results and rarely test stable demonstration use. \benchmark{} adds a stability track to the same shot-scaling settings: for each query, the task definition and metric stay fixed while the demonstrations are replaced, reordered, or mixed with noisy examples. The resulting drops measure whether gains rely on stable context use rather than a particular demonstration set or order.

\noindent\textbf{Benchmark split integrity.}
To prevent benchmark leakage, we curate \benchmark{} independently from the training episodes and audit each benchmark-aligned training pool. All tasks are exact-deduplicated with task-specific keys. Feature-Based tasks further use task-aware similarity filtering in the multimodal embedding space. Understanding and perceptual tasks use a 0.5 threshold, while generation-heavy tasks use a 0.6 threshold, following the practical duplicate-detection band reported for embedding-based contamination audits~\cite{spiesberger2026soft}. Intent-Based tasks are isolated by exact task-key and core-intent matching. The retained training-pool sizes are reported in the Supplementary Material. After filtering, Feature-Based pools remain below their thresholds even at the averaged per-query peak: the maxima are 0.494 for 0.5-threshold tasks and 0.595 for 0.6-threshold tasks, and the largest averaged distribution mean is 0.247. The audit statistics in \cref{tab:purity_similarity} support the integrity of the benchmark split.
\subsection{CAPM for Unified Multimodal ICL}
\label{sec:cari}

\subsubsection{Motivation, Setup, and Assumptions}
\label{sec:capm_setup}

\begin{figure*}[!t]
    \centering
    \includegraphics[width=\textwidth]{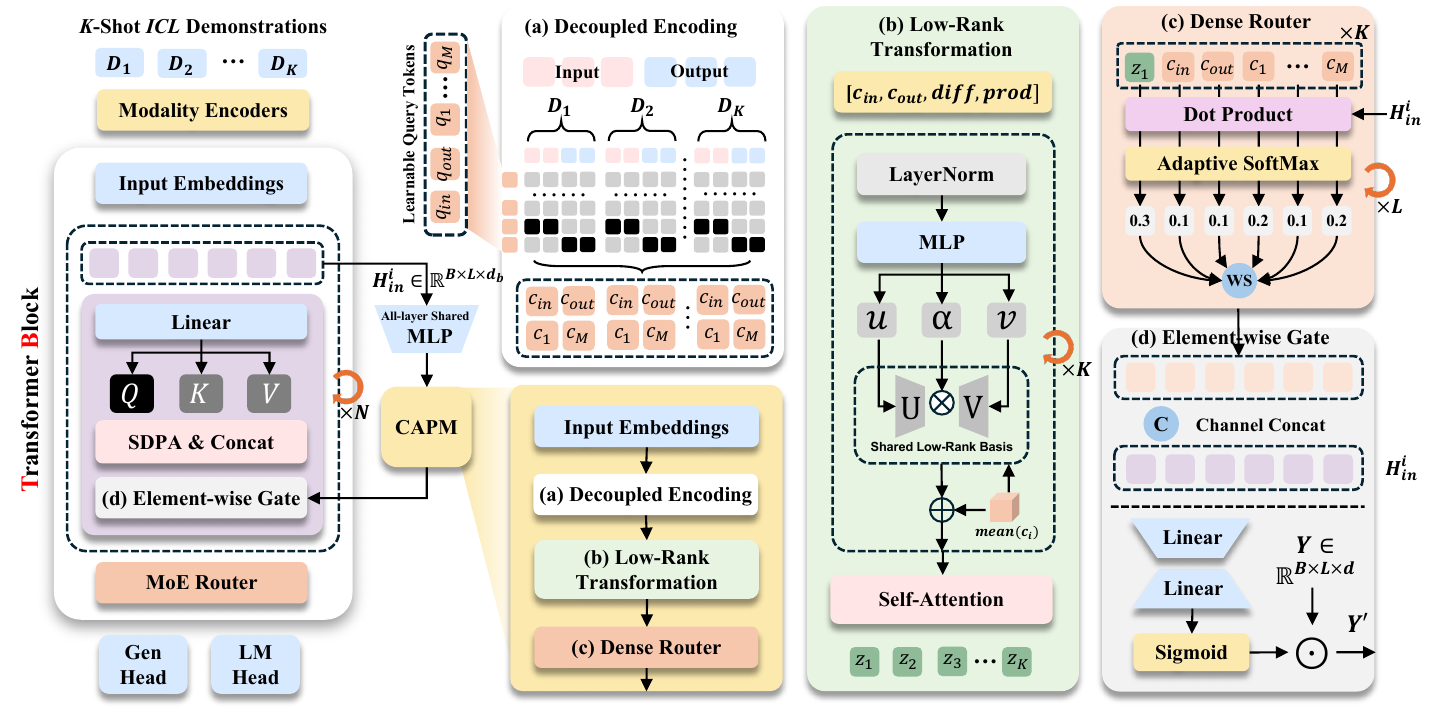}
    \caption{\textbf{CAPM adapts Transformer-based models through a four-stage context-modulation pipeline:} \textbf{\textit{(a)}} segment-masked decoupling constructs role-specific anchors, \textbf{\textit{(b)}} a rank-factorized descriptor compresses each demonstration into a transformation token, \textbf{\textit{(c)}} adaptive dense routing retrieves context with a task-conditioned temperature, and \textbf{\textit{(d)}} a near-identity element-wise gate re-injects this context multiplicatively.}
    \label{fig:CAPM}
\end{figure*}

The dataset and benchmark of \cref{sec:dataset} can expose \emph{where} unified ICL breaks down but not \emph{why} additional demonstrations can destabilize adaptation. We attribute this to a property of standard self-attention, which consumes a few-shot episode as a flat token stream. Three failure modes are relevant: \textit{(i) role conflation}, where user instructions and assistant responses mix before the input$\rightarrow$output mapping is isolated, \textit{(ii) context dilution}, where hundreds of interleaved tokens per shot make the useful transformation signal sparse, and \textit{(iii) static aggregation}, where a fixed routing sharpness is poorly matched to both retrieval-heavy and inductive episodes. These modes are consistent with the non-monotonic scaling we observe. To counter them, we introduce the \textbf{Context-Adaptive Prototype Modulator}, or CAPM, shown in \cref{fig:CAPM}. CAPM is a plug-and-play adapter that intercepts a subset of backbone layers and rewrites few-shot context through three operations matched to the three modes: segment-masked \emph{disentanglement} for role-specific anchors, \emph{rank-factorized transformation extraction} for a compact relation update, and \emph{adaptive dense routing} for task-conditioned retrieval. The result is re-injected through a near-identity gate that preserves the pre-trained prior at initialization.

Let $\mathcal{D}=\{d_i\}_{i=1}^{k}$ be the $k$-shot context of a UniICL episode, where each demonstration $d_i$ consists of a user instruction and an assistant response embedded by the backbone into $X_i \in \mathbb{R}^{L_i \times d_b}$. CAPM projects each $X_i$ into a working dimension $d_p$, constructs role-specific anchors and generic context slots, and encodes the input$\rightarrow$output relation through a rank-factorized transformation token. The resulting prototypes are aggregated into a context signal $C_t$ for each backbone state $h_t$ and fused with the scaled dot-product attention (SDPA) output through a multiplicative residual gate. Our construction rests on three assumptions that formalize the failure-mode analysis above and make the desired compression explicit.

\begin{assumption}[Role Segmentability]
For every $d_i$, the user-instruction and assistant-response spans are known. Masked probes restricted to the corresponding spans yield role-specific anchors $c_{\text{in}}^{(i)}$ and $c_{\text{out}}^{(i)}$ whose direct token access is disentangled, so the input$\rightarrow$output relation can be estimated without letting the response anchor be dominated by instruction tokens or vice versa.
\end{assumption}

\begin{assumption}[Low-Rank Transformation Manifold]
Although the token-level realization of a demonstration is high-dimensional, the rule-relevant change it conveys is locally well approximated by a rank-$r$ operator acting on a compact context token, with $r \ll d_p$. Thus a demonstration need not expose all $L_i d_b$ token coordinates to the backbone. It is sufficient to expose a low-rank transformation branch plus calibrated anchors for retrieval.
\end{assumption}

\begin{assumption}[Task-Conditional Routing Sharpness]
The optimal retrieval distribution over prototypes depends on both query-prototype similarity and episode-level context: perception-dominant episodes often demand a sharply peaked, low-temperature match to specific demonstrations, whereas inductive episodes benefit from broader, high-temperature aggregation across demonstrations. Because this cognitive demand is reflected in the demonstration set, routing sharpness should be inferred dynamically rather than fixed \emph{a priori}.
\end{assumption}
\begin{table}[t]

\centering
\caption{\textbf{Task-level similarity distribution after final filtering for retained Feature-Based tasks.} Each statistic is first computed per benchmark query and then averaged over queries.}
\label{tab:purity_similarity}
\footnotesize
\setlength{\tabcolsep}{3pt}
\renewcommand{\arraystretch}{1.15}
\begin{tabular*}{\columnwidth}{@{\extracolsep{\fill}}lcccccc@{}}
\toprule
\textbf{Task} & \(\boldsymbol{N_q}\) & \textbf{Mean\({}^{q}\)} & \textbf{Median\({}^{q}\)} & \textbf{P90\({}^{q}\)} & \textbf{Top10\({}^{q}\)} & \textbf{Peak\({}^{q}\)} \\
\midrule
VG & 100 & 0.020 & 0.015 & 0.072 & 0.440 & 0.481 \\
AR & 100 & 0.192 & 0.187 & 0.251 & 0.480 & 0.494 \\
SR & 100 & 0.139 & 0.134 & 0.200 & 0.459 & 0.484 \\
SAC & 50 & 0.023 & 0.016 & 0.082 & 0.457 & 0.485 \\
IG & 50 & 0.247 & 0.234 & 0.353 & 0.580 & 0.593 \\
IM & 50 & 0.236 & 0.228 & 0.311 & 0.583 & 0.595 \\
AA & 100 & 0.018 & 0.014 & 0.070 & 0.435 & 0.473 \\
FD & 100 & 0.030 & 0.019 & 0.107 & 0.446 & 0.474 \\
VR & 50 & 0.072 & 0.062 & 0.161 & 0.451 & 0.482 \\
\bottomrule
\end{tabular*}
\raggedright
\footnotesize Task names use initials, e.g., VG (Visual Grounding).
\end{table}

\noindent\textbf{Why does disentangled low-rank modulation stabilize scaling?}
The key observation is that CAPM restricts the explicit transformation branch through which a demonstration acts on the backbone. Whereas flat attention exposes up to $L_i d_b$ token coordinates to every later query, CAPM first summarizes the input$\rightarrow$output relation as a compact update to a generic context token. The next statement makes this bottleneck precise. It is the theoretical lens we use to interpret our ablations.

\begin{proposition}[Rank-Factorized Demonstration Modulation]
Let $a_k^{(i)} := U_{\text{base},k}\!\odot u^{(i)}_k$ and $b_k^{(i)} := V_{\text{base},k}\!\odot v^{(i)}_k$, and define the demonstration-specific operator
\[
    A^{(i)} = \eta \sum_{k=1}^{r}\alpha^{(i)}_k a_k^{(i)} {b_k^{(i)}}^\top .
\]
Stage~\textit{(b)} computes $\tilde z^{(i)} = g^{(i)} + A^{(i)}g^{(i)}$. Hence $\mathrm{rank}(A^{(i)}) \le r$, and the induced update $\delta^{(i)} := \tilde z^{(i)}-g^{(i)}$ lies in $\mathrm{Im}(A^{(i)})$, an at most $r$-dimensional subspace of $\mathbb{R}^{d_p}$ independent of the demonstration length $L_i$.
\end{proposition}

\noindent This follows because $A^{(i)}$ is a sum of $r$ rank-one outer products. The proposition characterizes the inductive bias imposed by CAPM: each demonstration's explicit transformation passes through a rank-$r$ operator independent of $L_i$, preserving the rule-relevant component under Assumption~2 while suppressing token-level redundancy. The remaining role anchors are retained as calibrated prototypes in the routing bank of stage~\textit{(c)}, so the backbone still accesses fine-grained demonstration content. Only the transformation channel is rank-restricted.

\par\vspace{0.15\baselineskip}
\noindent\textbf{Remark.} \textit{The proposition and assumptions predict which components are structurally necessary. Removing stage~\textit{(b)} deletes the rank-factorized transformation branch and should especially harm generation, which relies on explicit cross-modal mapping. Removing stage~\textit{(a)} weakens the role-specific anchors and should inflate context perturbation through role conflation. Fixing the routing temperature in stage~\textit{(c)} violates Assumption~3 and should suppress the learning curve. These predictions are corroborated by the ablations in \cref{tab:ablation_component,tab:ablation_tau}.}

\subsubsection{CAPM Architecture}
\label{sec:capm_architecture}

The four stages instantiate the above design.

\noindent\textbf{\textit{(a)} Decoupled Demonstration Encoding.}
To enforce the structural disentanglement required by Assumption~1, we introduce segment-masked cross-attention. Shared backbone embeddings $X_i \in \mathbb{R}^{L_i \times d_b}$ are projected to $d_p$, and $Y_i = \mathrm{CrossAttn}(Q, X_i W_{\text{in}}, M_i)$ is applied with learnable queries $Q = [q_{\text{in}}, q_{\text{out}}, q_1, \dots, q_K]$. The segment mask $M_i$ constrains $q_{\text{in}}$ and $q_{\text{out}}$ to attend exclusively to the user input and assistant response, while the remaining $K$ probes attend to the full sequence. This yields $[c_{\text{in}}^{(i)}, c_{\text{out}}^{(i)}, C^{(i)}]$, separating the instruction anchor, the response anchor, and generic context slots $C^{(i)} = [c_1^{(i)}, \dots, c_K^{(i)}]$. Restricting $q_{\text{in}}/q_{\text{out}}$ to their respective spans makes the anchors role-specific, so the downstream differential features capture the input$\rightarrow$output mapping rather than superficial token co-occurrence. As examined in \cref{tab:ablation_component}, this stage acts as the role-separation step that keeps the transformation descriptor from mixing prompt semantics with response content.

\noindent\textbf{\textit{(b)} Low-Rank Transformation and Interaction.}
To abstract the transformation induced by the demonstration onto the rank-$r$ manifold of Assumption~2, we derive a compact pre-interaction token $\tilde z^{(i)}$. We first pool the generic context slots into a global token, $g^{(i)} = \mathrm{Mean}\big(\mathrm{RMSNorm}(C^{(i)})\big)$. To capture transition dynamics, we compute the differential feature $\Delta^{(i)} = c_{\text{out}}^{(i)} - c_{\text{in}}^{(i)}$ and the interactive feature $\Pi^{(i)} = c_{\text{in}}^{(i)} \odot c_{\text{out}}^{(i)}$ between anchors. These are concatenated with the original anchors to form a relation descriptor $\phi^{(i)}$, which predicts dynamic modulation coefficients $u^{(i)}, v^{(i)} \in \mathbb{R}^{r \times d_p}$ and $\alpha^{(i)}\in\mathbb{R}^{r}$ for the predefined rank $r$:
\begin{align}
\phi^{(i)}
&= \mathrm{LayerNorm}\bigl(\mathrm{Concat}[c_{\text{in}}^{(i)},\, c_{\text{out}}^{(i)},\notag\\
&\hspace{4.6em}
\Delta^{(i)},\, \Pi^{(i)}]\bigr), \label{eq:caro_phi}\\
[u^{(i)}, v^{(i)}, \alpha^{(i)}]
&= \mathcal{H}_{\text{coef}}(\phi^{(i)}). \notag
\end{align}
To avoid instantiating a full $d_p \times d_p$ operator matrix per demonstration, which would re-introduce the redundancy Assumption~2 asks us to remove, we maintain shared global bases $U_{\text{base}}, V_{\text{base}} \in \mathbb{R}^{r \times d_p}$ and let the per-demonstration coefficients modulate them. The transformation token $\tilde z^{(i)}$ is obtained by applying the resulting rank-factorized operator to $g^{(i)}$:
\begin{align}
    a_k^{(i)} &= U_{\text{base}, k} \odot u^{(i)}_k,\qquad
    b_k^{(i)} = V_{\text{base}, k} \odot v^{(i)}_k, \notag\\
    A^{(i)} &= \eta \sum_{k=1}^{r}\alpha^{(i)}_k a_k^{(i)} {b_k^{(i)}}^\top, \label{eq:caro_lowrank}\\
    \tilde 
    z^{(i)} &= g^{(i)} + \eta \sum_{k=1}^r \alpha^{(i)}_k
    a_k^{(i)} \langle b_k^{(i)},\, g^{(i)} \rangle . \notag
\end{align}
Because the modulation rank $r$ is shared across demonstrations while the coefficients $(u^{(i)}, v^{(i)}, \alpha^{(i)})$ are instance-specific, $\tilde z^{(i)}$ exposes a compact transformation branch whose operator rank is bounded by $r$. Finally, inter-demonstration relationships are modeled by self-attention over $\tilde Z=[\tilde z^{(1)}, \dots, \tilde z^{(k)}]$, yielding $Z=[z^{(1)}, \dots, z^{(k)}]$ and allowing the model to reconcile or weight competing rules rather than treating demonstrations independently. As examined in \cref{tab:ablation_component}, this branch serves as the rule-compression step that turns each demonstration into a compact update while retaining the transferable input$\rightarrow$output relation.

\noindent\textbf{\textit{(c)} Adaptive Dense Routing.}
For each demonstration, we assemble a latent prototype bank $B^{(i)} = [z^{(i)}, c_{\text{in}}^{(i)}, c_{\text{out}}^{(i)}, C^{(i)}]$. Since these prototypes originate from different branches, each token $B_s$ is first calibrated by a type-wise affine transform,
\begin{equation}
    \mathcal{B}_s = \gamma_{\kappa_s}\odot B_s + \beta_{\kappa_s},
    \qquad \kappa_s\in\{z,c_{\text{in}},c_{\text{out}},C\},
\end{equation}
with $\gamma$ and $\beta$ initialized to one and zero. Backbone states query the calibrated bank $B_{\text{cal}}=\{\mathcal{B}_s\}_s$ via dense cosine routing. Crucially, and in direct response to Assumption~3, the routing temperature itself is not fixed but is dynamically inferred from the aggregated context $z_{\text{pool}} = \mathrm{Mean}(Z)$ by an MLP:
\begin{equation}
\begin{aligned}
    \delta_\tau &= s_\tau\tanh(\mathrm{MLP}_{\tau}(z_{\text{pool}})),\\
    \tau &= \tau_{\min} + (\tau_{\max}-\tau_{\min}) \cdot \sigma(\tau_0+\delta_\tau),
\end{aligned}
\end{equation}
\begin{equation}
\begin{aligned}
    \mathcal{A}_{t,s}
    &= \frac{\langle \overline{\psi(h_t)}, \overline{\mathcal{B}}_s \rangle}{\tau},\\
    C_t
    &= \sum_{s} \mathrm{softmax}(\mathcal{A}_{t,s}) \cdot \mathcal{B}_s.
\end{aligned}
\end{equation}
Here $\overline{\cdot}$ is $\ell_2$ normalization, making the score a cosine similarity. The offset scale $s_\tau$ is a hyper-parameter and is set to $0.25$ in all experiments. Because $\tau$ is read from the episode context rather than tuned globally, the model can sharpen its match toward $\tau_{\min}$ for retrieval-heavy episodes and broaden it toward $\tau_{\max}$ for inductive ones. The dynamic-$\tau$ ablation tests this adaptation at the task level. Dense routing, rather than top-$1$ routing, additionally preserves a soft residual over all prototypes, so informative but non-top demonstrations are not silently discarded. As examined in \cref{tab:ablation_component,tab:ablation_tau} and stress-tested in \cref{fig:robustness_curves,tab:robustness_table}, this stage acts as the context-selection step that decides which prototypes should modulate each query and how sharp that modulation should be.

\noindent\textbf{\textit{(d)} Element-wise Gating Injection.}
We first project the routed context back to the backbone width, $\bar C_t=C_tW_c\in\mathbb{R}^{d_b}$, and concatenate it with the normalized hidden state: $X_{\text{gate}} = [\mathrm{LayerNorm}(H_{\text{in}, t}), \bar C_t]$. The gating multiplier $m_t\in\mathbb{R}^{d_b}$ is computed by a bottleneck MLP with GELU:
\begin{equation}
    m_t = \sigma\bigl(W_2 \mathrm{GELU}(W_1 X_{\text{gate}} + b_1) + b_2\bigr),
\end{equation}
where $W_2$ is zero-initialized and $b_2=2.0$, so $m_t$ starts as the input-independent constant $\sigma(2.0)\approx0.88$. This positive-bias initialization gives a near-identity multiplier while still leaving room for modulation. Contextual injection is applied multiplicatively as $Y' = Y \odot m_t$, where $Y$ is the SDPA output~\cite{qiu2025gated}. The multiplicative form is deliberate: it lets context adjust per-channel gain on the attention output without overwriting the internal representations that unified understanding and generation both rely on. As examined in \cref{tab:ablation_component,fig:feature_control,fig:feature_attention}, this stage provides a safe injection path that redirects late-layer attention toward the selected context while preserving the backbone pathway. The overhead of this injection is reported in \cref{tab:inference_cost}.

\section{Experiments}
\label{sec:experiments}

In this section, we evaluate \method{} and use {\benchmark} to characterize unified multimodal ICL. The evaluation spans understanding and generation, comparing with state-of-the-art unified models and strong open-source MLLMs. Main results define the performance profile, taxonomy-level trends explain shot-scaling differences across capabilities, and VL-ICL-Bench tests transfer beyond our suite. Comprehensive ablations and diagnostics then attribute gains to curated data, CAPM stabilization, and metric design.
\subsection{Experimental Setup}
\noindent{\textbf{Baselines.}}
We select BAGEL~\cite{deng2025emerging} as our primary baseline. Beyond delivering competitive unified understanding and generation performance, BAGEL's Mixture-of-Transformers architecture enables bidirectional self-attention across the understanding and generation token spaces, thereby allowing CAPM to operate on both branches simultaneously. We compare against two groups. The first is state-of-the-art (SOTA) unified models: UniWorld-V1~\cite{lin2025uniworld}, Nexus-Gen-V2~\cite{zhang2025nexus}, and Ovis-U1~\cite{wang2025ovis}. The second is strong open-source MLLMs at various parameter scales: Qwen3-VL~\cite{bai2025qwen3}, Qwen2.5-VL~\cite{qwen2.5-VL}, and InternVL-3.5~\cite{wang2025internvl3}. We include MLLMs because recent unified models have begun matching them on static understanding benchmarks, making ICL a sharper and more diagnostic test of genuine generalization capability.

\noindent{\textbf{Training Strategy.}}
To stabilize optimization and disentangle understanding and generation dynamics, we employ a two-stage training paradigm on a single H20 GPU node. Both stages use a maximum learning rate of $2 \times 10^{-5}$ with a 500-step linear warm-up, followed by a constant schedule. \textbf{Stage-I: Understanding Warm-up.} We first establish robust in-context adaptation by optimizing solely the understanding ICL objective for 10,000 steps with a maximum sequence length of 20,480 tokens, while freezing the generation head. \textbf{Stage-II: Unified ICL Training.} Shifting the primary focus to generation ICL, we interleave a small proportion of understanding data to mitigate catastrophic forgetting. The overall loss combines CE and MSE with a 1:1 fixed ratio. This stage trains for 10,000 steps with an expanded 28,672-token maximum sequence length. \looseness=-1

\begin{table*}[!t]
    \centering
    \caption{\textbf{Aggregated main results across the six capability categories.} For each category, \textbf{Und.} averages understanding-side primary metrics and \textbf{Gen.} reports the category's generation-side metric. Each model is summarized by \textbf{Z.s.} for zero-shot, \textbf{Pk.} for peak over 0/1/2/4/8-shot, and \textbf{Eff.} for ICL efficiency. Z.s./Pk. are on a normalized 0--100 scale. A zero efficiency means the shot curve is flat relative to zero-shot, while a negative value means additional demonstrations reduce the average score. \method{} gives the strongest average peak and efficiency, and is the only unified model with positive efficiency on both sides.}
    \label{tab:main_results_summary}
    {
    \footnotesize
    \setlength{\tabcolsep}{2.8pt}
    \renewcommand{\arraystretch}{1.2}
    \newcommand{\sotahighlight}[1]{\textbf{\textcolor{eccvblue}{#1}}}
    \newcommand{\effhighlight}[1]{\textbf{\textcolor{red}{#1}}}
    \resizebox{\textwidth}{!}{%
    \begin{tabular}{c c c c c c c c c c c c c c c c}
    \toprule
    \multirow{2}{*}{Model} & \multirow{2}{*}{Stat.} & \multicolumn{2}{c}{Perception} & \multicolumn{2}{c}{Imitation} & \multicolumn{2}{c}{Concept.} & \multicolumn{2}{c}{Deduction} & \multicolumn{2}{c}{Analogy} & \multicolumn{2}{c}{Discern.} & \multicolumn{2}{c}{Average} \\
    \cmidrule(lr){3-4} \cmidrule(lr){5-6} \cmidrule(lr){7-8} \cmidrule(lr){9-10} \cmidrule(lr){11-12} \cmidrule(lr){13-14} \cmidrule(lr){15-16}
     &  & Und. & Gen. & Und. & Gen. & Und. & Gen. & Und. & Gen. & Und. & Gen. & Und. & Gen. & Und. & Gen. \\
    \hline
    \rowcolor{gray_tab}
    \multicolumn{16}{c}{\textbf{Unified Models}} \\
    \hline
     & Z.s. & 31.1 & 83.5 & 69.4 & 71.4 & 23.0 & 61.9 & 69.0 & 53.1 & 41.0 & 58.6 & 55.3 & 14.7 & 48.1 & 57.2 \\
    UniWorld-V1~\cite{lin2025uniworld} & Pk. & 38.8 & 83.5 & 73.9 & 76.2 & 43.0 & 61.9 & 69.0 & 53.1 & 74.1 & 58.6 & 55.3 & 51.1 & 53.9 & 57.2 \\
     & Eff. & 2.9 & -25.3 & 3.6 & \effhighlight{2.8} & 7.8 & -12.3 & -- & -- & 19.9 & -12.1 & -8.2 & 24.3 & 1.3 & -3.8 \\
    \hline
     & Z.s. & 39.1 & 76.4 & 71.3 & 57.1 & 24.0 & 55.3 & 51.0 & 52.5 & 20.7 & 67.1 & 47.6 & 68.5 & 42.3 & 62.8 \\
    Nexus-Gen-V2~\cite{zhang2025nexus} & Pk. & 39.1 & 76.4 & 71.3 & 60.8 & 33.0 & 55.3 & 51.0 & 52.5 & 21.0 & 67.1 & 63.3 & 78.7 & 42.3 & 62.8 \\
     & Eff. & -8.8 & \effhighlight{-9.6} & -11.1 & 0.9 & 5.7 & -7.2 & -- & -- & 0.1 & -11.6 & 10.4 & 8.9 & -2.4 & -1.8 \\
    \hline
     & Z.s. & 54.8 & 85.7 & 72.1 & 68.9 & 22.0 & 60.3 & 36.0 & 58.2 & 36.9 & 57.3 & 62.7 & 86.4 & 47.4 & 69.5 \\
    Ovis-U1~\cite{wang2025ovis} & Pk. & 54.8 & 85.7 & 72.1 & 68.9 & 31.0 & 62.9 & 36.0 & 58.2 & 37.8 & 57.3 & 62.7 & \sotahighlight{86.4} & 47.4 & 69.5 \\
     & Eff. & -37.2 & -18.1 & -20.9 & -0.4 & 4.4 & -7.0 & -- & -- & -1.5 & -4.4 & -21.5 & -7.9 & -13.2 & -5.5 \\
    \hline
     & Z.s. & 72.7 & 84.4 & 69.3 & 77.0 & 23.0 & 60.5 & 60.0 & 53.8 & 40.6 & 65.3 & 51.6 & 8.1 & 52.8 & 58.2 \\
    BAGEL~\cite{deng2025emerging} & Pk. & 72.7 & 84.4 & 72.8 & 77.0 & 41.0 & 72.4 & 60.0 & 53.8 & 68.6 & 65.3 & 66.4 & 45.3 & 59.3 & 60.5 \\
     & Eff. & -14.9 & -20.3 & 2.2 & -6.6 & 11.9 & 10.8 & -- & -- & 22.3 & -15.4 & 6.7 & 26.1 & 4.3 & -0.3 \\
    \hline
    \rowcolor{eccvblue!10}  & Z.s. & 66.9 & \begin{tabular}[c]{@{}c@{}}86.5\\[-2pt]{\tiny\textcolor{green!60!black}{$\uparrow$2.0}}\end{tabular} & 60.8 & \begin{tabular}[c]{@{}c@{}}78.8\\[-2pt]{\tiny\textcolor{green!60!black}{$\uparrow$1.8}}\end{tabular} & 20.0 & \begin{tabular}[c]{@{}c@{}}62.2\\[-2pt]{\tiny\textcolor{green!60!black}{$\uparrow$1.7}}\end{tabular} & \begin{tabular}[c]{@{}c@{}}88.0\\[-2pt]{\tiny\textcolor{green!60!black}{$\uparrow$28.0}}\end{tabular} & \begin{tabular}[c]{@{}c@{}}62.2\\[-2pt]{\tiny\textcolor{green!60!black}{$\uparrow$8.4}}\end{tabular} & 37.0 & \begin{tabular}[c]{@{}c@{}}68.0\\[-2pt]{\tiny\textcolor{green!60!black}{$\uparrow$2.7}}\end{tabular} & \begin{tabular}[c]{@{}c@{}}83.1\\[-2pt]{\tiny\textcolor{green!60!black}{$\uparrow$31.5}}\end{tabular} & \begin{tabular}[c]{@{}c@{}}22.7\\[-2pt]{\tiny\textcolor{green!60!black}{$\uparrow$14.7}}\end{tabular} & \begin{tabular}[c]{@{}c@{}}59.3\\[-2pt]{\tiny\textcolor{green!60!black}{$\uparrow$6.5}}\end{tabular} & \begin{tabular}[c]{@{}c@{}}63.4\\[-2pt]{\tiny\textcolor{green!60!black}{$\uparrow$5.2}}\end{tabular} \\
    \rowcolor{eccvblue!10} UniICL (Ours) & Pk. & \begin{tabular}[c]{@{}c@{}}\sotahighlight{80.9}\\[-2pt]{\tiny\textcolor{green!60!black}{$\uparrow$8.2}}\end{tabular} & \begin{tabular}[c]{@{}c@{}}\sotahighlight{86.5}\\[-2pt]{\tiny\textcolor{green!60!black}{$\uparrow$2.0}}\end{tabular} & \begin{tabular}[c]{@{}c@{}}76.6\\[-2pt]{\tiny\textcolor{green!60!black}{$\uparrow$3.8}}\end{tabular} & \begin{tabular}[c]{@{}c@{}}\sotahighlight{79.3}\\[-2pt]{\tiny\textcolor{green!60!black}{$\uparrow$2.3}}\end{tabular} & \begin{tabular}[c]{@{}c@{}}\sotahighlight{70.0}\\[-2pt]{\tiny\textcolor{green!60!black}{$\uparrow$29.0}}\end{tabular} & \begin{tabular}[c]{@{}c@{}}\sotahighlight{78.8}\\[-2pt]{\tiny\textcolor{green!60!black}{$\uparrow$6.4}}\end{tabular} & \begin{tabular}[c]{@{}c@{}}\sotahighlight{88.0}\\[-2pt]{\tiny\textcolor{green!60!black}{$\uparrow$28.0}}\end{tabular} & \begin{tabular}[c]{@{}c@{}}\sotahighlight{62.2}\\[-2pt]{\tiny\textcolor{green!60!black}{$\uparrow$8.4}}\end{tabular} & \begin{tabular}[c]{@{}c@{}}\sotahighlight{85.4}\\[-2pt]{\tiny\textcolor{green!60!black}{$\uparrow$16.8}}\end{tabular} & \begin{tabular}[c]{@{}c@{}}\sotahighlight{74.1}\\[-2pt]{\tiny\textcolor{green!60!black}{$\uparrow$8.8}}\end{tabular} & \begin{tabular}[c]{@{}c@{}}\sotahighlight{87.3}\\[-2pt]{\tiny\textcolor{green!60!black}{$\uparrow$20.9}}\end{tabular} & \begin{tabular}[c]{@{}c@{}}60.9\\[-2pt]{\tiny\textcolor{green!60!black}{$\uparrow$15.6}}\end{tabular} & \begin{tabular}[c]{@{}c@{}}\sotahighlight{78.9}\\[-2pt]{\tiny\textcolor{green!60!black}{$\uparrow$19.7}}\end{tabular} & \begin{tabular}[c]{@{}c@{}}\sotahighlight{69.6}\\[-2pt]{\tiny\textcolor{green!60!black}{$\uparrow$9.2}}\end{tabular} \\
    \rowcolor{eccvblue!10}  & Eff. & \begin{tabular}[c]{@{}c@{}}\effhighlight{9.7}\\[-2pt]{\tiny\textcolor{green!60!black}{$\uparrow$24.6}}\end{tabular} & \begin{tabular}[c]{@{}c@{}}-14.1\\[-2pt]{\tiny\textcolor{green!60!black}{$\uparrow$6.3}}\end{tabular} & \begin{tabular}[c]{@{}c@{}}\effhighlight{13.9}\\[-2pt]{\tiny\textcolor{green!60!black}{$\uparrow$11.6}}\end{tabular} & \begin{tabular}[c]{@{}c@{}}-0.2\\[-2pt]{\tiny\textcolor{green!60!black}{$\uparrow$6.4}}\end{tabular} & \begin{tabular}[c]{@{}c@{}}\effhighlight{39.8}\\[-2pt]{\tiny\textcolor{green!60!black}{$\uparrow$27.9}}\end{tabular} & \begin{tabular}[c]{@{}c@{}}\effhighlight{11.5}\\[-2pt]{\tiny\textcolor{green!60!black}{$\uparrow$0.7}}\end{tabular} & -- & -- & \begin{tabular}[c]{@{}c@{}}\effhighlight{44.7}\\[-2pt]{\tiny\textcolor{green!60!black}{$\uparrow$22.4}}\end{tabular} & \begin{tabular}[c]{@{}c@{}}\effhighlight{-1.7}\\[-2pt]{\tiny\textcolor{green!60!black}{$\uparrow$13.8}}\end{tabular} & 3.1 & \begin{tabular}[c]{@{}c@{}}\effhighlight{28.0}\\[-2pt]{\tiny\textcolor{green!60!black}{$\uparrow$1.9}}\end{tabular} & \begin{tabular}[c]{@{}c@{}}\effhighlight{16.9}\\[-2pt]{\tiny\textcolor{green!60!black}{$\uparrow$12.5}}\end{tabular} & \begin{tabular}[c]{@{}c@{}}\effhighlight{4.9}\\[-2pt]{\tiny\textcolor{green!60!black}{$\uparrow$5.2}}\end{tabular} \\
    \hline
    \rowcolor{gray_tab}
    \multicolumn{16}{c}{\textbf{MLLMs}} \\
    \hline
     & Z.s. & 80.5 & -- & 79.5 & -- & 24.0 & -- & 83.0 & -- & 40.9 & -- & 57.6 & -- & 60.9 & -- \\
    Qwen3-VL-8B~\cite{bai2025qwen3} & Pk. & 80.5 & -- & 81.6 & -- & 66.0 & -- & 83.0 & -- & 84.4 & -- & 69.4 & -- & 73.5 & -- \\
     & Eff. & -4.8 & -- & 1.3 & -- & 28.9 & -- & -- & -- & 37.7 & -- & 5.1 & -- & 9.5 & -- \\
    \hline
     & Z.s. & 79.7 & -- & 83.8 & -- & 24.0 & -- & 82.0 & -- & 41.3 & -- & 68.8 & -- & 63.3 & -- \\
    Qwen3-VL-32B~\cite{bai2025qwen3} & Pk. & 80.1 & -- & \sotahighlight{85.1} & -- & 63.0 & -- & 82.0 & -- & 83.4 & -- & 68.8 & -- & 75.6 & -- \\
     & Eff. & 0.2 & -- & 0.4 & -- & 18.9 & -- & -- & -- & 35.5 & -- & -2.5 & -- & 7.0 & -- \\
    \hline
     & Z.s. & 68.1 & -- & 74.7 & -- & 22.0 & -- & 63.0 & -- & 38.2 & -- & 45.6 & -- & 51.9 & -- \\
    Qwen2.5-VL-7B~\cite{qwen2.5-VL} & Pk. & 68.1 & -- & 76.1 & -- & 38.0 & -- & 63.0 & -- & 70.4 & -- & 45.6 & -- & 53.5 & -- \\
     & Eff. & -12.5 & -- & 1.0 & -- & 9.1 & -- & -- & -- & 22.7 & -- & -8.6 & -- & 0.3 & -- \\
    \hline
     & Z.s. & 72.2 & -- & 72.9 & -- & 24.0 & -- & 68.0 & -- & 38.2 & -- & 65.8 & -- & 56.8 & -- \\
    Qwen2.5-VL-32B~\cite{qwen2.5-VL} & Pk. & 72.2 & -- & 79.2 & -- & 38.0 & -- & 68.0 & -- & 81.5 & -- & 71.7 & -- & 68.5 & -- \\
     & Eff. & -0.7 & -- & 4.6 & -- & 5.5 & -- & -- & -- & 35.8 & -- & -5.1 & -- & 6.0 & -- \\
    \hline
     & Z.s. & 41.0 & -- & 74.7 & -- & 21.0 & -- & 80.0 & -- & 38.0 & -- & 38.5 & -- & 48.9 & -- \\
    InternVL3.5-8B~\cite{wang2025internvl3} & Pk. & 54.1 & -- & 76.8 & -- & 24.0 & -- & 80.0 & -- & 55.2 & -- & 47.0 & -- & 49.5 & -- \\
     & Eff. & 9.5 & -- & 0.9 & -- & 0.7 & -- & -- & -- & 13.0 & -- & -3.4 & -- & -1.7 & -- \\
    \hline
     & Z.s. & 48.4 & -- & 77.9 & -- & 22.0 & -- & 77.0 & -- & 41.5 & -- & 42.0 & -- & 51.5 & -- \\
    InternVL3.5-38B~\cite{wang2025internvl3} & Pk. & 50.0 & -- & 79.2 & -- & 27.0 & -- & 77.0 & -- & 73.7 & -- & 64.1 & -- & 58.7 & -- \\
     & Eff. & 1.1 & -- & 0.7 & -- & 3.4 & -- & -- & -- & 24.0 & -- & \effhighlight{15.3} & -- & 4.1 & -- \\
    \bottomrule
    \end{tabular}%
    }
    }
\end{table*}

\begin{figure*}[t]
    \centering
    \includegraphics[width=\textwidth]{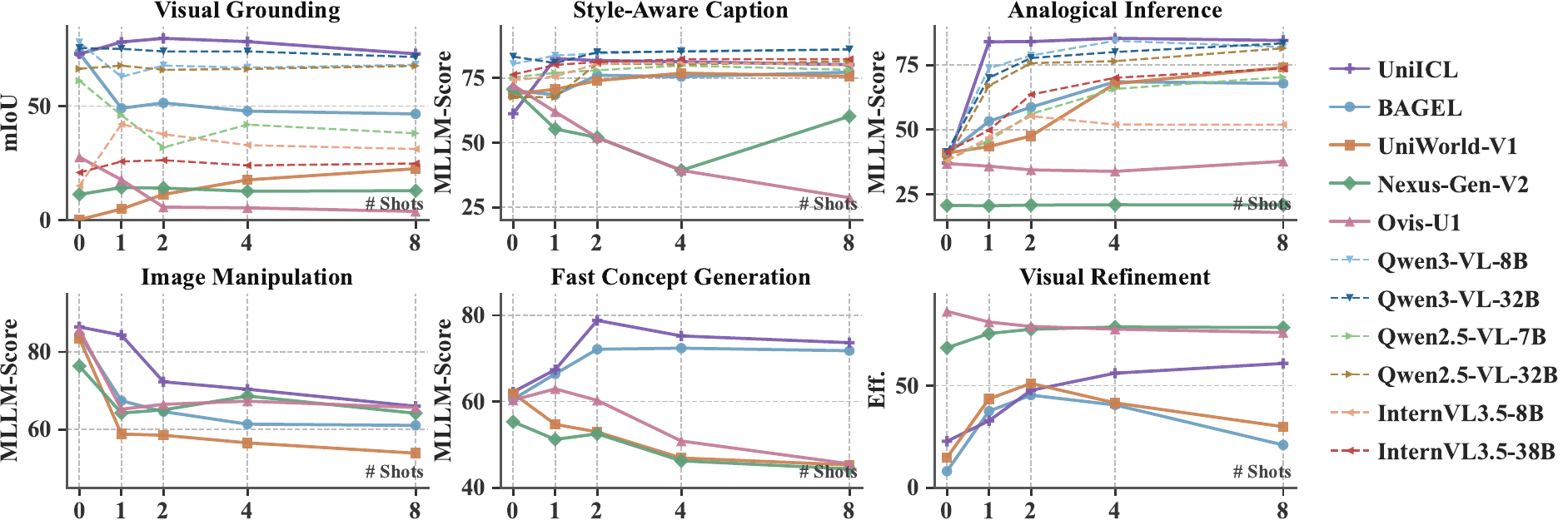}
    \caption{\textbf{k-shot performance curves for a subset of tasks.}}
    \label{fig:main_icl_curves}
\end{figure*}
\subsection{Main Results on {\benchmark}}

We summarize performance across the six capability categories in \cref{tab:main_results_summary,fig:main_icl_curves}. Peak performance alone hides how a model uses its demonstrations, so we also report an ICL efficiency metric: the normalized area of the performance delta relative to the zero-shot baseline under the $k$-shot curve.
\begin{equation}
\begin{aligned}
\text{Eff.}
&= \frac{1}{K_{\max}} \sum_{i=1}^{n} \Delta_i,\\
\Delta_i
&= \frac{(P_{k_i} - P_0) + (P_{k_{i-1}} - P_0)}{2}
(k_i - k_{i-1}),
\end{aligned}
\end{equation}
where $k_i \in \{0, 1, 2, 4, 8\}$, $K_{max}=8$, and $P_0$ is the zero-shot performance. A model can reach a high peak yet remain inefficient, gaining little from additional shots or even losing accuracy as the context grows, so we read peak and efficiency together rather than separately. Complete per-shot task-level scores are provided in the Supplementary Material.

\subsubsection{General ICL phenomena.}
\textbf{\textit{(1) Unification taxes ICL stability.}} Specialized MLLMs scale more gracefully on understanding. Qwen3-VL-32B rises from $63.3$ to $75.6$ between $0$- and $8$-shot with efficiency $+7.0$, and Qwen3-VL-8B behaves similarly at $+9.5$. Unified models pay a steeper few-shot cost. Ovis-U1 understanding drops monotonically from $47.4$ to $31.5$ despite a respectable zero-shot floor, UniWorld-V1 generation falls by about $9$ points, Nexus-Gen-V2 stays nearly flat with negative efficiency on both sides, and three of the four unified peers show negative efficiency on at least one side. The cost appears even when the zero-shot score is high: Ovis-U1 starts at $69.5$ generation but slips below $62$ by $8$-shot. This is consistent with cross-modal interference in the shared token space rather than weak base capability, and it is the behavior our method is designed to counter.
\textbf{\textit{(2) Understanding and generation asymmetry.}} Generation ICL is uniformly weaker than understanding across every model. Generation efficiency hovers near zero or turns negative even for strong models, with BAGEL at $-0.3$, UniWorld-V1 at $-3.8$, and Nexus-Gen-V2 at $-1.8$, while understanding efficiency stays positive on the same models. The gap is structural: generating an output from demonstrations likely requires stricter contextual grounding than selecting among given answers, since the model must reconstruct rather than merely align. This is why we evaluate generative stability separately rather than folding it into a single score.
\textbf{\textit{(3) Cognitive level governs scaling behavior.}} The taxonomy tracks how demonstrations act. Perception-level tasks, with strong zero-shot priors, often \emph{degrade} under additional shots, whereas higher levels such as Conception and Analogy gain more. We test this by correlating shot-trend vectors across pairs of tasks and report the category-level matrix in \cref{fig:taxonomy_trend_corr}. Aggregating the underlying task-pair correlations gives $0.746$ within the same taxonomy level against $0.063$ across levels. Deduction is handled separately because its demonstrations form a causal chain, so its within-level consistency is set to $1$ by construction. The gap between $0.746$ and $0.063$ indicates that the six levels capture distinct scaling regimes rather than a single broad cluster. The Perception and Imitation pair is a useful test case: their correlation is $0.62$ since both attend to visual evidence, yet their curves diverge. In our model Perception peaks early at $80.9$ at $2$-shot and then falls to $74.0$ at $8$-shot, while Imitation climbs steadily from $60.8$ to $76.6$. The split is sharper in baselines, where Ovis-U1 Perception collapses from $54.8$ to $6.9$ while Imitation degrades more gently. This moderate rather than perfect correlation supports treating them as separate capability levels.

\subsubsection{Results of \method{}.}
\textbf{\textit{(1) A balanced, ICL-positive unified profile.}} \method{} attains the highest average peak on both understanding and generation among unified models, $78.9$ and $69.6$, and also achieves the highest efficiency on both sides, $+16.9$ and $+4.9$. It is the \emph{only} unified model with positive efficiency on both understanding and generation, whereas the other unified peers are negative or near zero on at least one side. On understanding peak, \method{} also matches or surpasses the specialized MLLMs, $78.9$ against $75.6$ for Qwen3-VL-32B, while retaining the generation ability those models lack entirely.
\textbf{\textit{(2) Sample-efficient scaling.}} \method{} reaches its strongest understanding result at just $2$ shots and remains close through $4$ shots, showing that the gain comes from effective use of a small context rather than from simply adding more demonstrations. The raw BAGEL backbone behaves differently: it peaks later on understanding and then declines without stabilizing, while its generation curve is nearly flat. Specialized MLLMs also remain below our $2$-shot understanding score even at their best reported shot setting, with Qwen3-VL-32B reaching $75.6$. Thus the main point is not monotonic shot growth. It is that \method{} extracts useful context earlier and more reliably, despite starting from a backbone whose own $8$-shot peak sits below $58$.
\textbf{\textit{(3) Gain sources: data leads, CAPM stabilizes.}} The dominant capability driver is {\dataset}. Compared with the unadapted BAGEL score of $59.3/60.5$, the w/o CAPM variant already reaches $77.0/68.6$ in \cref{tab:ablation_component}, so most of the peak gain comes from the data and training rather than from the module. CAPM mainly stabilizes few-shot adaptation, raising generation ICL efficiency from $0.4$ to $4.9$. Data thus establishes capability while CAPM delivers stable scaling, so we attribute the peak gain to the data rather than to the module. The largest categorical jump is also where one would expect demonstration use to matter most, Conception from $41.0$ for BAGEL to $70.0$, a level that relies on binding novel concepts from examples.
\textbf{\textit{(4) Uneven gains reveal the remaining bottlenecks.}} The strongest ICL gains appear where examples carry transferable structure: Conception and Analogy reach the highest understanding efficiencies, $39.8$ and $44.7$, while Discernment shows the strongest generation gains at $28.0$ efficiency. High-prior visual generation is more vulnerable to context overload: Perception generation is still negative at $-14.1$ and Imitation generation is nearly flat at $-0.2$. The clearest failure mode is fine-grained \emph{Image Manipulation}, where dense visual contexts strain cross-modal alignment and in turn cap \emph{Analogical Editing} at higher shots. We illustrate this behavior in \cref{fig:bad_case}. \method{} still degrades more gently than the unified baselines on these regimes, and resolving the residual bottleneck is left to future work.

\noindent\begin{minipage}{\columnwidth}
    \centering
    \captionsetup{hypcap=false}
    \includegraphics[width=\columnwidth]{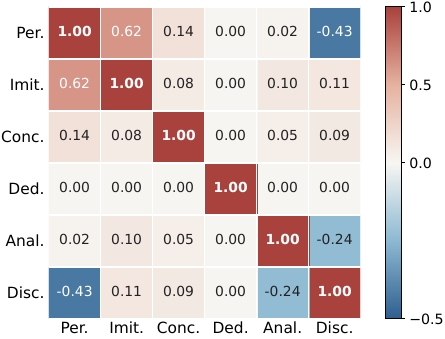}
    \captionof{figure}{\textbf{Taxonomy shot-trend correlation matrix.} Values denote Spearman correlations between category-level shot-scaling trends.}
    \label{fig:taxonomy_trend_corr}
\end{minipage}\par

\begin{table*}[t]
    \centering
    \caption{\textbf{Cross-benchmark validation on VL-ICL-Bench understanding tasks.} \method{} improves the BAGEL backbone and matches GPT-4V on average peak and efficiency. Zero-shot values can be \(0.0\) for tasks that depend on demonstrations, and negative Eff. indicates that additional shots lower the average score relative to zero-shot.}
    \label{tab:vlicl_understanding}
    \footnotesize
    \setlength{\tabcolsep}{2.0pt}
    \renewcommand{\arraystretch}{1.1}
    \resizebox{\textwidth}{!}{%
    \begin{tabular}{@{}c*{21}{c}@{}}
        \toprule
        \multirow{2}{*}{Model}
        & \multicolumn{3}{c}{\shortstack{Fast Open-Ended\\MiniImageNet}}
        & \multicolumn{3}{c}{\shortstack{CLEVR\\Count Induction}}
        & \multicolumn{3}{c}{\shortstack{Operator\\Induction}}
        & \multicolumn{3}{c}{\raisebox{4pt}{TextOCR}}
        & \multicolumn{3}{c}{\shortstack{Interleaved\\Operator Induction}}
        & \multicolumn{3}{c}{\shortstack{Fast Matching\\MiniImageNet}}
        & \multicolumn{3}{c}{\raisebox{4pt}{Avg.}} \\
        \cmidrule(lr){2-4}\cmidrule(lr){5-7}\cmidrule(lr){8-10}\cmidrule(lr){11-13}\cmidrule(lr){14-16}\cmidrule(lr){17-19}\cmidrule(lr){20-22}
        & Z.s. & Pk. & Eff.
        & Z.s. & Pk. & Eff.
        & Z.s. & Pk. & Eff.
        & Z.s. & Pk. & Eff.
        & Z.s. & Pk. & Eff.
        & Z.s. & Pk. & Eff.
        & Z.s. & Pk. & Eff. \\
        \midrule
        BAGEL~\cite{deng2025emerging}
        & 0.0 & 72.0 & 51.4
        & 0.0 & 60.5 & 49.7
        & 0.0 & 25.0 & 10.0
        & 13.5 & 56.5 & \textbf{39.1}
        & \textbf{38.3} & 65.0 & 13.3
        & 0.0 & 56.2 & 46.2
        & 8.6 & 55.9 & 29.4 \\
        GPT4V~\cite{wu2024gpt}
        & 0.0 & 78.0 & 41.8
        & \textbf{6.0} & 42.0 & 29.0
        & 24.0 & \textbf{92.0} & \textbf{59.0}
        & 39.3 & 50.0 & 7.2
        & 36.0 & \textbf{74.0} & \textbf{32.2}
        & 0.0 & 58.2 & 38.3
        & 17.6 & 65.7 & 34.6 \\
        LLaVA-OV~\cite{li2024llava}
        & 0.0 & \textbf{98.7} & \textbf{68.4}
        & 0.5 & 42.3 & 32.7
        & 33.3 & 75.6 & 29.8
        & 48.5 & 51.7 & 2.4
        & \textbf{38.3} & 47.8 & 2.8
        & 0.0 & \textbf{70.3} & \textbf{50.8}
        & 20.1 & 64.4 & 31.1 \\
        VILA~\cite{lin2023vila}
        & 0.0 & 38.2 & 32.3
        & 3.5 & 34.3 & 27.5
        & 28.3 & 28.3 & \mbox{-18.9}
        & 28.0 & 30.2 & \mbox{-3.7}
        & 28.3 & 28.3 & \mbox{-15.7}
        & 0.0 & 49.9 & 44.6
        & 14.7 & 34.9 & 11.0 \\
        \midrule\noalign{\vskip 1pt}
        \rowcolor{eccvblue!10} UniICL (Ours)
        & 0.0 & 83.5 & 60.7
        & 0.0 & \textbf{62.5} & \textbf{50.7}
        & \textbf{38.3} & 81.7 & 28.8
        & \textbf{53.5} & \textbf{58.5} & 3.2
        & 30.0 & 55.0 & 18.9
        & 0.0 & 57.2 & 47.1
        & \textbf{20.3} & \textbf{66.4} & \textbf{34.9} \\
        \bottomrule
    \end{tabular}%
    }
\end{table*}

\begin{table*}[t]
    \centering
    \caption{\textbf{Cross-benchmark validation on VL-ICL-Bench generation tasks.} \method{} achieves the best average peak and efficiency among the open baselines.}
    \label{tab:vlicl_generation}
    \footnotesize
    \setlength{\tabcolsep}{3.2pt}
    \renewcommand{\arraystretch}{1.1}
    \begin{tabularx}{\textwidth}{@{}>{\centering\arraybackslash}p{3.1cm}*{15}{>{\centering\arraybackslash}X}@{}}
        \toprule
        \multirow{2}{*}{Model}
        & \multicolumn{3}{c}{\shortstack{Text-to-Image\\Fast MiniImageNet}}
        & \multicolumn{3}{c}{\raisebox{4pt}{CoBSAT}}
        & \multicolumn{3}{c}{\raisebox{4pt}{Fast Counting}}
        & \multicolumn{3}{c}{\shortstack{Fast Attribute\\Matching}}
        & \multicolumn{3}{c}{\raisebox{4pt}{Avg.}} \\
        \cmidrule(lr){2-4}\cmidrule(lr){5-7}\cmidrule(lr){8-10}\cmidrule(lr){11-13}\cmidrule(lr){14-16}
        & Z.s. & Pk. & Eff.
        & Z.s. & Pk. & Eff.
        & Z.s. & Pk. & Eff.
        & Z.s. & Pk. & Eff.
        & Z.s. & Pk. & Eff. \\
        \midrule
        BAGEL~\cite{deng2025emerging}
        & 0.5 & 34.5 & 26.7
        & 6.5 & 50.5 & 38.5
        & \textbf{70.0} & 75.0 & 1.5
        & 18.0 & 33.5 & 11.9
        & \textbf{23.8} & 48.4 & 19.6 \\
        SEED-LLaMA-14B~\cite{ge2023planting}
        & 0.8 & 21.2 & 16.3
        & 5.5 & 43.8 & 30.7
        & 41.3 & 51.6 & 1.9
        & 22.4 & 35.6 & 8.4
        & 17.5 & 38.1 & 14.3 \\
        Emu2-Gen~\cite{sun2024generative}
        & 0.0 & \textbf{37.0} & \textbf{28.6}
        & \textbf{8.7} & 28.7 & 15.6
        & 44.2 & 59.2 & 5.2
        & \textbf{26.5} & 34.0 & -0.2
        & 19.9 & 39.7 & 12.3 \\
        \midrule\noalign{\vskip 1pt}
        \rowcolor{eccvblue!10} UniICL (Ours)
        & \textbf{1.0} & 36.5 & \textbf{28.6}
        & 6.5 & \textbf{63.0} & \textbf{49.5}
        & 62.5 & \textbf{80.0} & \textbf{9.3}
        & 17.5 & \textbf{42.5} & \textbf{19.5}
        & 21.9 & \textbf{55.5} & \textbf{26.7} \\
        \bottomrule
    \end{tabularx}
\end{table*}

\subsection{Cross-Benchmark Generalization}
To test transfer beyond \benchmark{}, we evaluate \method{} on VL-ICL-Bench~\cite{zong2024vl} under the original benchmark protocol, including the same task split, shot settings, and zero-shot/peak-shot/ICL-efficiency metrics. For external baselines, we report representative models and values from the VL-ICL-Bench paper in \cref{tab:vlicl_understanding,tab:vlicl_generation} rather than newly re-run results, covering closed-source, open-source MLLM, and generation-oriented baselines such as GPT-4V~\cite{wu2024gpt}, LLaVA-OneVision-72B~\cite{li2024llava}, VILA-7B~\cite{lin2023vila}, SEED-LLaMA-14B~\cite{ge2023planting}, Emu2-Gen~\cite{sun2024generative}, and BAGEL~\cite{deng2025emerging}. Under this matched protocol, \method{} improves the BAGEL backbone on both sides. On understanding, zero-shot rises from $8.6$ to $20.3$ and peak from $55.9$ to $66.4$, with efficiency moving from $29.4$ to $34.9$. The peak and efficiency land close to GPT-4V, $66.4$ versus $65.7$ and $34.9$ versus $34.6$, from a much smaller open backbone. On generation, peak rises from $48.4$ to $55.5$ and efficiency from $19.6$ to $26.7$. This external benchmark indicates that the learned demonstration assembly and CAPM stabilization transfer beyond the suite used to develop our taxonomy. The understanding-side lift is also larger than the generation-side lift in relative terms, which matches the asymmetry observed on \benchmark{} and suggests the same bottleneck carries over.

\begin{figure*}[!t]
    \centering
    \includegraphics[width=\textwidth]{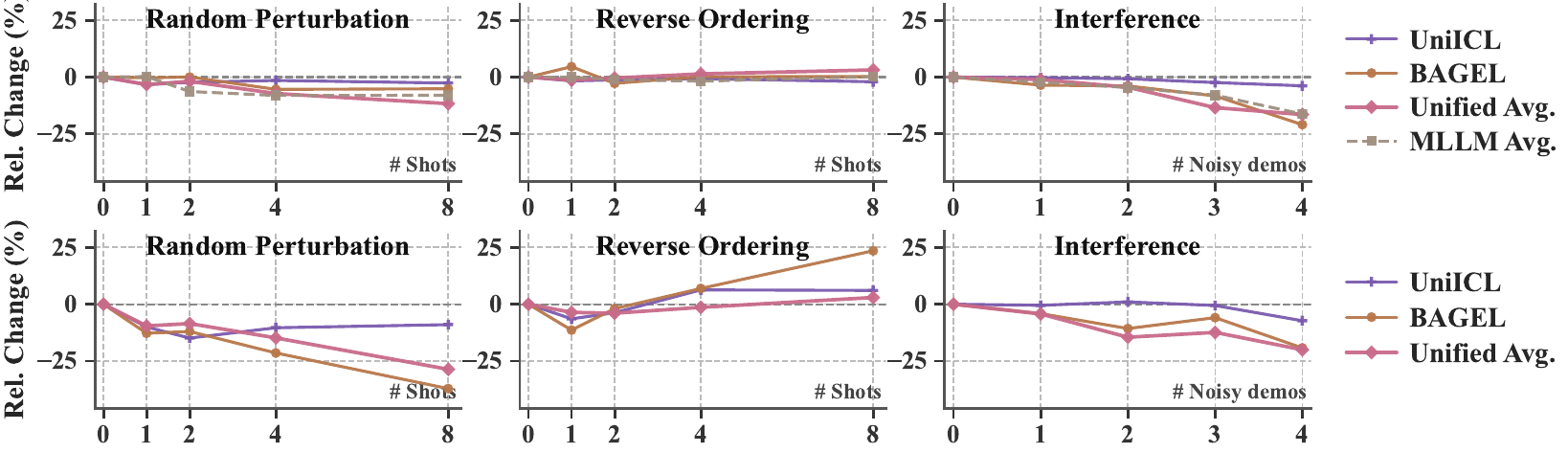}
    \caption{\textbf{Signed changes under context perturbations.} Random replacement and interference mostly lower scores, while reverse ordering tests order sensitivity.}
    \label{fig:robustness_curves}
\end{figure*}
\begin{table}[t]
    \centering
    \caption{\textbf{ICL stability comparison.} Values are absolute relative changes. Lower is more stable.}
    \label{tab:robustness_table}
    \footnotesize
    \setlength{\tabcolsep}{1.5pt}
    \renewcommand{\arraystretch}{1.3}
    \begin{tabularx}{\columnwidth}{@{}l*{4}{>{\centering\arraybackslash}X}@{}}
        \toprule
        \multicolumn{1}{c}{Model} & \makecell{Random\\Replace} & \makecell{Reverse\\Order} & Interference & Average \\
        \midrule
        \rowcolor{gray_tab}\multicolumn{5}{c}{\textbf{Understanding}} \\
        \midrule
        MLLM Avg.    & $7.3\%$  & \underline{$1.8\%$} & \underline{$6.3\%$}  & \underline{$5.1\%$} \\
        Unified Avg. & $17.2\%$ & $8.5\%$ & $11.3\%$ & $12.4\%$ \\
        BAGEL        & \underline{$7.1\%$}  & $2.8\%$ & $7.9\%$  & $5.9\%$ \\
        \rowcolor{eccvblue!10}
        Ours         & \textbf{2.1\%} & \textbf{1.4\%} & \textbf{1.6\%} & \textbf{1.7\%} \\
        \midrule
        \rowcolor{gray_tab}\multicolumn{5}{c}{\textbf{Generation}} \\
        \midrule
        Unified Avg. & \underline{$15.7\%$} & \textbf{5.7\%}  & $10.4\%$ & \underline{$10.6\%$} \\
        BAGEL        & $22.0\%$ & $10.9\%$ & \underline{$7.8\%$}  & $13.6\%$ \\
        \rowcolor{eccvblue!10}
        Ours         & \textbf{10.3\%} & \underline{$6.1\%$} & \textbf{3.4\%} & \textbf{6.6\%} \\
        \bottomrule
    \end{tabularx}
\end{table}

\subsection{Stability Analysis}

We evaluate random replacement, reverse ordering, and interference in \cref{fig:robustness_curves,tab:robustness_table}. The signed curves show that random replacement and interference mostly lower performance, supporting the assembly design: demonstrations cannot be organized arbitrarily. For reverse ordering, a score may rise or fall after reordering, but either sign still reflects sensitivity to arbitrary order. The curves in \cref{fig:robustness_curves} keep signed group-level trends to show direction, whereas \cref{tab:robustness_table} reports scalar stability after removing signed cancellation. The signed point values are listed in the Supplementary Material. For each model and metric \(m\), we compute the absolute relative change under perturbation family \(\xi\), average over metrics \(\mathcal{M}\), and integrate the variation curve:
\begin{equation}
\begin{aligned}
d_{\xi}(r_j)
&= \frac{100}{|\mathcal{M}|}\sum_{m\in\mathcal{M}}
\left|\frac{P_{\xi,m}(r_j)-P_m(r_j)}{P_m(r_j)}\right|,\\
\mathcal{A}_{\xi}
&= \frac{1}{r_T-r_0}\sum_{j=1}^{T}
\frac{d_{\xi}(r_j)+d_{\xi}(r_{j-1})}{2}\Delta r_j .
\end{aligned}
\label{eq:stability}
\end{equation}
Here \(r_j\) is the perturbation index, \(P_m(r_j)\) and \(P_{\xi,m}(r_j)\) are the clean and perturbed scores, and \(\Delta r_j=r_j-r_{j-1}\). For random replacement and reverse ordering, \(r_j\) is the shot count with \(r_j\in\{0,1,2,4,8\}\), \(d_\xi(0)=0\), and \(r_T-r_0=8\). For interference, \(r_j\) is the noise level from the clean four-shot setting to level \(4\), so \(r_T-r_0=4\). For average rows, we first compute \(\mathcal{A}_\xi\) for each model and then average the resulting scores. \method{} is the most stable model overall, with average variation of only $1.7\%$ on understanding and $6.6\%$ on generation, against $12.4\%$ and $10.6\%$ for the unified average. Four patterns emerge from the curves.
\textbf{\textit{(1) Demonstration identity is the main sensitivity.}} Random replacement is the largest source of variation for the unified-model average, $17.2\%$ on understanding and $15.7\%$ on generation, and the signed curves show that it mainly causes performance drops. \method{} separates earliest from the pack, holding to $2.1\%$ and $10.3\%$. Matched demonstrations therefore matter more than context length: more shots do not help if the shots are wrong. This sensitivity motivates the assembly ablation in \cref{tab:assembly_ablation}.
\textbf{\textit{(2) Demonstration identity dominates ordering.}} Reverse ordering is the mildest perturbation for every model, with the unified average shifting only $8.5\%$ on understanding and $5.7\%$ on generation, and several models moving within $2\%$. Some signed curves improve after reversal, but this is not reliable progress. It means the model is sensitive to presentation order. Unified ICL therefore appears primarily content-driven: assembled-set compatibility matters more than demonstration order.
\textbf{\textit{(3) The clearest stabilizer-specific gain appears under interference.}} Interference is where adaptive routing should matter most, and \method{} separates most clearly there: absolute variation is only $1.6\%$/$3.4\%$ on understanding/generation, against $7.9\%$/$7.8\%$ for BAGEL and $11.3\%$/$10.4\%$ for the unified average. The signed trend with severity is also telling. As mismatched demonstrations grow from $1$ to $4$, BAGEL drops by $-3.7\%$ to $-20.4\%$, a more than fivefold widening, whereas \method{} bends from $-7.6\%$ to $-13.7\%$ and stays roughly where the baseline starts. The module's benefit therefore scales with how noisy the context is, which is consistent with CAPM filtering perturbed context rather than helping uniformly.
\textbf{\textit{(4) Generation is the fragile side.}} Absolute variation is larger on generation than on understanding under both random replacement and interference. For BAGEL under random replacement the gap is $7.1\%$ on understanding versus $22.0\%$ on generation, roughly a threefold difference. The unified average differs because understanding-side perturbations vary more across models and tasks, so signed curves partially cancel while scalar scores do not. The pattern still tracks the main asymmetry: generating coherent outputs likely requires stricter contextual grounding, so a noisy demonstration corrupts the whole output rather than a single answer choice. This is why we weight generative stability in unified ICL evaluation.

\subsection{Qualitative Analysis}

\begin{figure*}[!t]
    \centering
    \includegraphics[width=\textwidth]{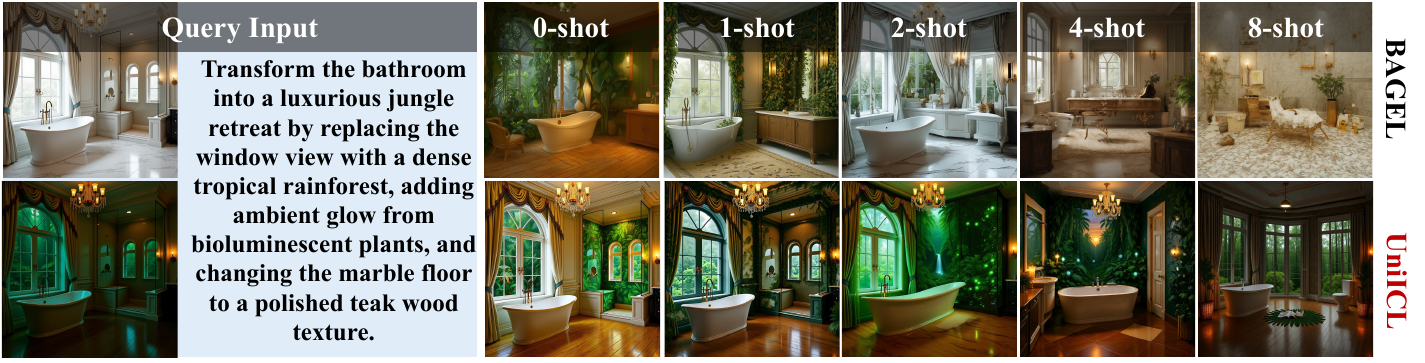}
    \caption{\textbf{Failure case on multi-shot image manipulation.} \method{} captures the intended edit and degrades less severely than the baselines, but dense pixel-level details still weaken localization.}
    \label{fig:bad_case}
\end{figure*}

\begin{figure}[t]
    \centering
    \includegraphics[width=\columnwidth]{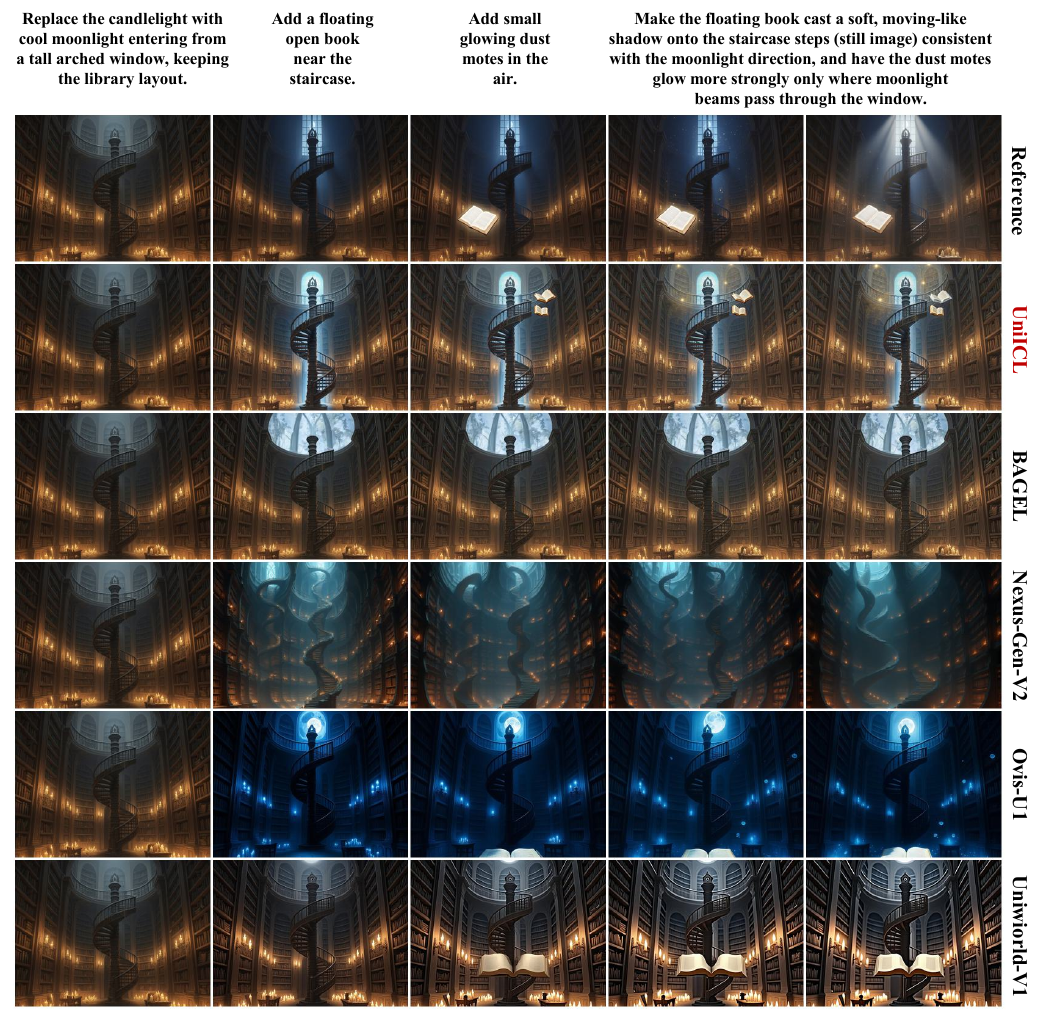}
    \caption{\textbf{Chain-of-Editing qualitative comparison.}}
    \label{fig:coe}
\end{figure}

We compare generation outputs under the $2$-shot setting across generation-side tasks. The pattern is consistent: baselines either miss the demonstrated transformation rule or produce object distortion in longer contexts, while \method{} applies the structural edit implied by the demonstrations and preserves non-target regions. On the benchmark-only \emph{Chain-of-Editing}, shown in \cref{fig:coe}, \method{} performs better by inferring the step-to-step editing rule and maintaining cumulative consistency. Nevertheless, \method{} does not fully solve fine-grained \emph{Image Manipulation} at higher shot counts. As shown in \cref{fig:bad_case}, given overlapping spatial transformations, \method{} captures the correct editing intent but produces imprecise localization. Even in this failure mode, its degradation is milder than that of baselines, since it preserves more of the intended edit while baseline outputs more often drift to wrong transformations or object distortion. This suggests that the failure is less about intent recognition than execution: the intended edit is identified, but dense pixel-level constraints dominate the final synthesis and lower pixel-sensitive scores. This is consistent with Image Manipulation having the lowest generation efficiency among all subtasks.

\begin{figure*}[!t]
    \centering
    \includegraphics[width=\textwidth]{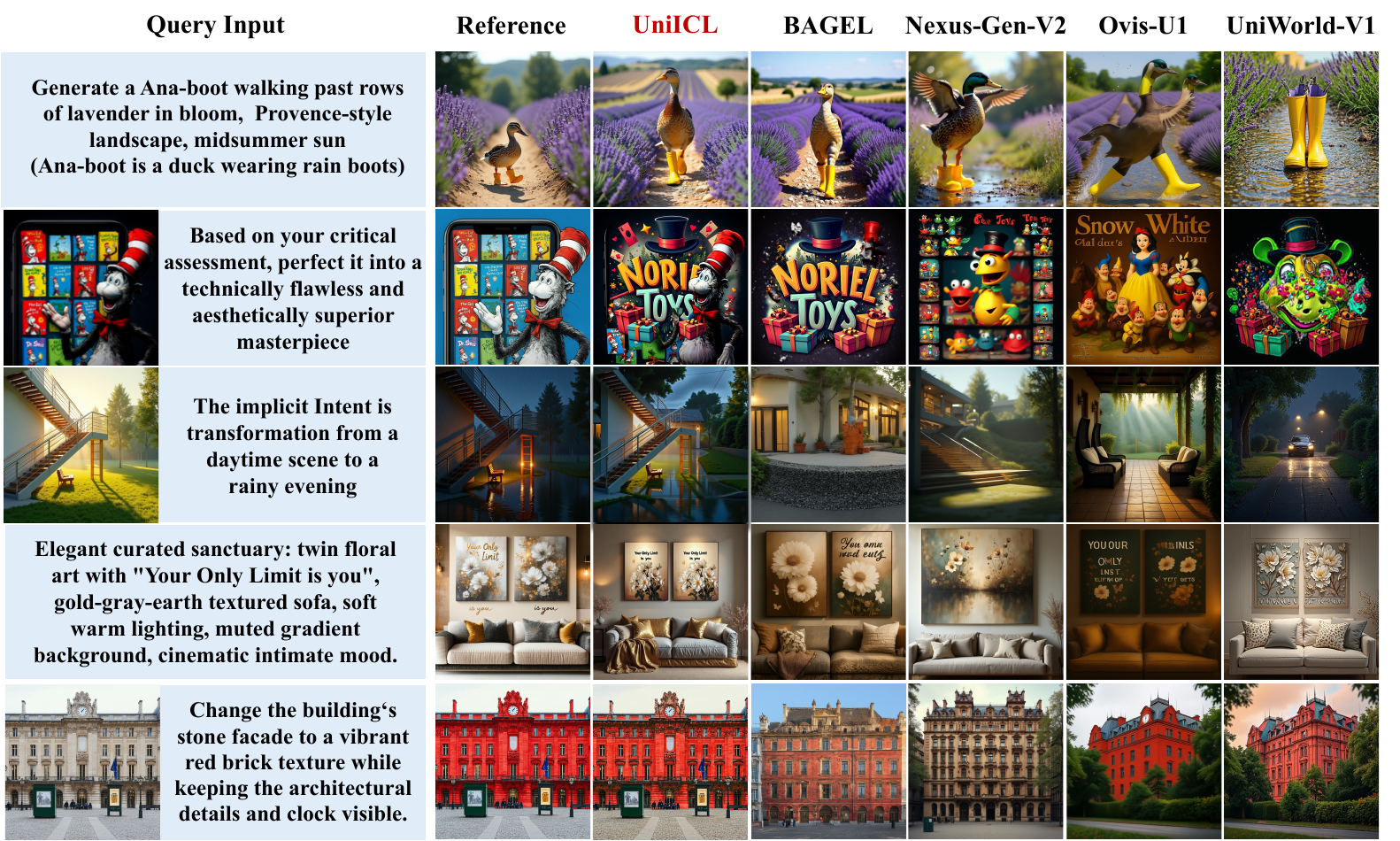}
    \caption{\textbf{Qualitative comparison of generative ICL tasks.} From top to bottom: Fast Concept Generation, Visual Refinement, Analogical Editing, Instructional Generation, and Image Manipulation.}
    \label{fig:case}
\end{figure*}

\begin{table}[t]
    \centering
    \caption{\textbf{Feature-Based demonstration assembly ablation.} The \(\Delta\) columns report changes relative to the default cosine+DPP rule (ours), and the row without similarity or diversification uses random selection.}
    \label{tab:assembly_ablation}
    \footnotesize
    \setlength{\tabcolsep}{2.7pt}
    \renewcommand{\arraystretch}{1.2}
    \renewcommand{\tabularxcolumn}[1]{m{#1}}
    \begin{tabularx}{\columnwidth}{@{}>{\centering\arraybackslash}X>{\centering\arraybackslash}Xcccc@{}}
        \toprule
        \multicolumn{2}{c}{Setting} & \multicolumn{2}{c}{Understanding} & \multicolumn{2}{c}{Generation} \\
        \cmidrule(lr){1-2}\cmidrule(lr){3-4}\cmidrule(lr){5-6}
        Similarity & Diversify & $\Delta$Peak & $\Delta$Eff. & $\Delta$Peak & $\Delta$Eff. \\
        \midrule
        \rowcolor{gray_tab}\multicolumn{6}{c}{\textbf{Demonstration assembly}} \\
        \xmark & \xmark & -1.45 & -2.09 & -13.34 & -15.17 \\
        Cosine & \xmark & -0.93 & -1.18 & -2.74 & -3.36 \\
        Cosine & K-means & -0.41 & \underline{-0.32} & -0.58 & -0.77 \\
        Cosine & MMR & \underline{-0.27} & -0.48 & -0.74 & \underline{-0.59} \\
        Euclidean & DPP & -0.35 & -0.66 & \underline{-0.49} & -1.04 \\
        \rowcolor{eccvblue!10}Cosine & DPP & \textbf{0.00} & \textbf{0.00} & \textbf{0.00} & \textbf{0.00} \\
        \bottomrule
    \end{tabularx}
\end{table}

\begin{table}[!t]

    \centering
    \caption{\textbf{CAPM ablation.} We report peak performance over shot settings and ICL efficiency.}
    \label{tab:combined_ablations}
    \label{tab:ablation_1}
    \label{tab:ablation_component}
    \label{tab:ablation_tau}
    \label{tab:ablation_2}
    \footnotesize
    \setlength{\tabcolsep}{2.7pt}
    \renewcommand{\arraystretch}{1.2}
    \renewcommand{\tabularxcolumn}[1]{m{#1}}%
    \begin{tabularx}{\columnwidth}{@{}>{\centering\arraybackslash}Xcccc@{}}
        \toprule
        & \multicolumn{2}{c}{Understanding} & \multicolumn{2}{c}{Generation} \\
        \cmidrule(lr){2-3}\cmidrule(lr){4-5}
        \multirow{-2}{*}{Setting} & Peak & ICL Eff. & Peak & ICL Eff. \\
        \midrule
        \rowcolor{gray_tab}\multicolumn{5}{c}{\textbf{CAPM component}} \\
        w/ Gate-only         & 77.6 & 15.5 & 69.0 & 1.8 \\
        w/o Adapt. Routing & \textbf{78.9} & \underline{16.4} & \textbf{69.8} & \underline{4.2} \\
        w/o Low-Rank Trans. & \underline{78.6} & \textbf{16.9} & 68.3 & 2.6 \\
        w/o Decoupled Enc.   & 78.0 & 15.7 & 68.8 & 3.0 \\
        w/o CAPM             & 77.0 & 15.1 & 68.6 & 0.4 \\
        \rowcolor{eccvblue!10}w/ CAPM & \textbf{78.9} & \textbf{16.9} & \underline{69.6} & \textbf{4.9} \\
        \rowcolor{gray_tab}\multicolumn{5}{c}{\textbf{Routing temperature}} \\
        Fixed $\tau=0.1$ & \textbf{79.0} & \underline{16.4} & 68.4 & 2.8 \\
        Fixed $\tau=0.4$ & 78.2 & 15.9 & \underline{68.9} & \underline{3.8} \\
        Fixed $\tau=0.7$ & 78.6 & 16.1 & 67.9 & 3.6 \\
        \rowcolor{eccvblue!10}Dynamic $\tau$ & \underline{78.9} & \textbf{16.9} & \textbf{69.6} & \textbf{4.9} \\
        \rowcolor{gray_tab}\multicolumn{5}{c}{\textbf{Injection depth}} \\
        7  & 76.1 & \underline{14.4} & \textbf{70.3} & \underline{3.7} \\
        14 & \underline{76.3} & 13.1 & 69.0 & 3.4 \\
        \rowcolor{eccvblue!10}28 & \textbf{78.9} & \textbf{16.9} & \underline{69.6} & \textbf{4.9} \\
        \bottomrule
    \end{tabularx}

\end{table}

\subsection{Ablation Study and Analysis}
\subsubsection{Demonstration Assembly}

This ablation tests whether the retrieval side of {\method} depends on a particular diversity heuristic or on the broader principle of selecting relevant and non-redundant demonstrations. We isolate the Feature-Based assembly branch in \cref{tab:assembly_ablation} and report deltas relative to the cosine+DPP rule. Intent-Based tasks keep their deterministic matching because their demonstrations are defined by task structure rather than by feature retrieval.
Changing the diversity rule only causes small movement. K-means, MMR, and Euclidean-DPP stay within roughly one point on peak score, and their efficiency losses remain modest on both output sides. This indicates that the assembly pipeline is not tuned to a fragile optimizer. The important factor is preserving relevance and coverage at the same time. Removing diversification under cosine retrieval already hurts generation more than understanding, with a $2.74$ peak drop and a $3.36$ efficiency drop. Fully random selection is much worse, reducing generation peak by $13.34$ and generation efficiency by $15.17$, while understanding drops by only $1.45$ and $2.09$. This asymmetry matches the stability results: wrong demonstrations corrupt how a generative output is constructed, whereas understanding often still benefits from the query and answer prior. The assembly rule is therefore not only a retrieval convenience. It determines whether the context supplies complementary evidence instead of repeated or mismatched examples.

\subsubsection{CAPM Component Contributions}
\textbf{\textit{(1) CAPM mainly stabilizes context use.}} The component ablation in \cref{tab:ablation_component} asks what CAPM adds beyond the capability already learned from {\dataset}. The w/o CAPM row remains strong, reaching $77.0/68.6$ peak on understanding and generation. This confirms that the data and unified training account for most of the raw task competence. CAPM improves peak scores to $78.9/69.6$, but the clearer effect is on shot scaling: generation efficiency rises from $0.4$ to $4.9$. CAPM is therefore best understood as a mechanism that makes learned ICL capability persist across the shot curve.
\textbf{\textit{(2) Adaptive routing improves consistency rather than raw peak.}} The gate-only variant improves over w/o CAPM but reaches only $1.8$ generation efficiency, so a learned multiplicative gate is insufficient. Removing adaptive routing still gives a high generation peak of $69.8$, but its efficiency remains below the full CAPM setting, $4.2$ versus $4.9$. The peak result shows that the model can still find a strong isolated context, while the efficiency gap indicates that routing mainly smooths how demonstrations remain useful as the number and mixture of examples change.
\textbf{\textit{(3) Transformation encoding protects the generation side.}} Removing the low-rank transformation or the decoupled encoding keeps generation peaks competitive, but reduces generation efficiency to $2.6$ and $3.0$. These two CAPM components encode the demonstrated input-to-output relation and separate instruction-side evidence from response-side evidence. Without either component, the model can still benefit on some settings, but generation scales less reliably because output construction depends on preserving the transformation expressed by the demonstrations. This component pattern supports the same interpretation as the main results: {\dataset} builds the unified ICL capability, and CAPM makes that capability more stable under longer and noisier contexts.

\subsubsection{CAPM Routing Temperature}
The routing-temperature sweep in \cref{tab:ablation_tau} tests whether CAPM should use one global routing sharpness for all tasks. A fixed temperature is too coarse for the taxonomy because perception-style queries often prefer a sharp nearest match, while analogy and refinement tasks need broader aggregation across demonstrations. Predicting $\tau$ from the pooled context gives the best efficiency on both sides, $16.9$ for understanding and $4.9$ for generation, and also gives the best generation peak, $69.6$.
The fixed settings reveal the trade-off. A sharp value, $\tau{=}0.1$, slightly improves understanding peak to $79.0$, but generation efficiency falls to $2.8$. A moderate value, $\tau{=}0.4$, recovers part of the generation efficiency but remains below dynamic routing. A broad value, $\tau{=}0.7$, weakens generation peak to $67.9$. The main loss from a fixed temperature is therefore not always visible in the best shot. It appears in the area under the shot curve, where a single routing regime cannot stay appropriate as tasks and context sizes change.

\subsubsection{CAPM Injection Depth}
The injection-depth sweep in \cref{tab:ablation_2} tests where CAPM should interact with the language stack. Full $28$-layer injection gives the best overall balance. Compared with $14$ layers, it improves understanding peak from $76.3$ to $78.9$, understanding efficiency from $13.1$ to $16.9$, and generation efficiency from $3.4$ to $4.9$. The $7$-layer variant reaches the highest generation peak, $70.3$, but its understanding peak drops to $76.1$ and its generation efficiency remains below the full setting.
The depth trend suggests that shallow CAPM injection can create a strong best-shot generation point but does not stabilize the unified model across shots. Late layers are where task-specific representations and cross-modal evidence are consolidated, so routing only in early layers leaves a stronger trade-off between understanding and generation. We use full-depth CAPM because the target behavior is not a single high peak. The target is stable ICL improvement across both output sides.

\subsubsection{Mutual Impact Between Branches}
This analysis tests whether understanding and generation should be trained as separate ICL problems. We compare single-branch objectives in \cref{tab:branch_transfer_detail} and evaluate each one on the opposite side. The Gen-only variant transfers positively to understanding on average, $+2.1\%$, with large gains on Perception and Discernment. It still hurts Conception by $-7.9\%$, which suggests that generative supervision alone does not preserve the fast symbol-binding behavior needed for concept tasks.
The Und-only variant shows the complementary pattern. It improves generation by $+3.6\%$ on average and gives the largest gain on Discernment, $+19.8\%$, where criteria and preference structure transfer naturally from understanding supervision. At the same time, it weakens Perception and Imitation on the generation side. The branch transfer is therefore selective rather than uniformly positive. Joint training is useful because each branch supplies constraints the other branch lacks, and the remaining failures identify where unified ICL still needs stronger cross-modal grounding.

\subsubsection{CAPM Inference Cost}
The cost sweep in \cref{tab:inference_cost} measures whether the stability gain from CAPM comes with a large deployment cost. We evaluate Visual Grounding on the Top-10 case-study set while sweeping \(K\in\{0,1,2,4,8\}\) shots and \(N\in\{0,7,14,28\}\) injection layers. A separate model is instantiated per depth so shallow variants are not underestimated.
The dominant cost is the multimodal context itself. Without CAPM, latency grows from \(1.32\)s at zero-shot to \(7.60\)s at eight-shot, and memory rises from \(28.11\) to \(30.96\)GiB. Full-depth CAPM adds a smaller increment on top of this long-context cost. At eight-shot, memory increases by \(0.36\)GiB and latency increases by about \(0.65\)s. The overhead also grows with shot count, which is expected because more demonstrations create more routed context. The parameter footprint follows the same conclusion: full CAPM has \(189.2\)M parameters, only \(1.29\%\) of the \(14.61\)B backbone, while the $7$- and $14$-layer variants contain \(60.9\)M and \(103.7\)M parameters. CAPM therefore improves few-shot stability with overhead dominated by the existing long-context forward pass rather than by the module itself.
\begin{table}[t]
    \centering
    \caption{\textbf{Branch transfer per-category relative change (\%) vs.\ BAGEL.} Gen$\rightarrow$Und is the Gen-only variant on understanding tasks. Und$\rightarrow$Gen is the Und-only variant.}
    \label{tab:branch_transfer_detail}
    \footnotesize
    \setlength{\tabcolsep}{1.2pt}
    \renewcommand{\arraystretch}{1.0}
    \begin{tabular*}{\columnwidth}{@{\extracolsep{\fill}}lccccccc@{}}
        \toprule
        Setting & Perc. & Imit. & Conc. & Dedu. & Anal. & Disc. & Avg. \\
        \midrule
        Gen$\rightarrow$Und. & $+7.7$  & $+0.8$  & $-7.9$  & $+1.7$  & $-0.6$  & $+11.0$ & $+2.1$ \\
        Und$\rightarrow$Gen. & $-2.5$  & $-1.4$  & $+2.7$  & $-0.5$  & $+3.3$  & $+19.8$ & $+3.6$ \\
        \bottomrule
    \end{tabular*}
\end{table}

\begin{table}[t]
    \centering
    \caption{\textbf{Inference cost of CAPM.} Cell shading encodes relative magnitude within each block (darker = larger cost).}
    \label{tab:inference_cost}
    \footnotesize
    \setlength{\tabcolsep}{6.0pt}
    \renewcommand{\arraystretch}{1.3}
    \begin{tabular}{c *{5}{c}}
        \toprule
        CAPM layer \(N\) & \(K{=}0\) & \(K{=}1\) & \(K{=}2\) & \(K{=}4\) & \(K{=}8\) \\
        \midrule
        \rowcolor{gray_tab}\multicolumn{6}{c}{\textbf{Inference latency (s)}} \\
        \midrule
        0         & \cellcolor{eccvblue!1}1.32 & \cellcolor{eccvblue!7}1.99 & \cellcolor{eccvblue!13}2.63 & \cellcolor{eccvblue!26}4.00 & \cellcolor{eccvblue!59}7.60 \\
        7         & \cellcolor{eccvblue!1}1.36 & \cellcolor{eccvblue!9}2.23 & \cellcolor{eccvblue!16}2.95 & \cellcolor{eccvblue!31}4.60 & \cellcolor{eccvblue!70}8.73 \\
        14        & 1.25 & \cellcolor{eccvblue!8}2.10 & \cellcolor{eccvblue!14}2.75 & \cellcolor{eccvblue!29}4.32 & \cellcolor{eccvblue!64}8.05 \\
        28        & 1.27 & \cellcolor{eccvblue!9}2.17 & \cellcolor{eccvblue!15}2.86 & \cellcolor{eccvblue!30}4.43 & \cellcolor{eccvblue!66}8.25 \\
        \midrule
        \rowcolor{gray_tab}\multicolumn{6}{c}{\textbf{Peak GPU memory (GiB)}} \\
        \midrule
        0         & 28.11 & \cellcolor{eccvblue!6}28.38 & \cellcolor{eccvblue!11}28.62 & \cellcolor{eccvblue!27}29.37 & \cellcolor{eccvblue!62}30.96 \\
        7         & \cellcolor{eccvblue!2}28.22 & \cellcolor{eccvblue!9}28.54 & \cellcolor{eccvblue!14}28.75 & \cellcolor{eccvblue!30}29.49 & \cellcolor{eccvblue!65}31.08 \\
        14        & \cellcolor{eccvblue!4}28.30 & \cellcolor{eccvblue!11}28.62 & \cellcolor{eccvblue!16}28.83 & \cellcolor{eccvblue!32}29.57 & \cellcolor{eccvblue!67}31.16 \\
        28        & \cellcolor{eccvblue!8}28.46 & \cellcolor{eccvblue!14}28.77 & \cellcolor{eccvblue!19}28.99 & \cellcolor{eccvblue!35}29.73 & \cellcolor{eccvblue!70}31.32 \\
        \bottomrule
    \end{tabular}
\end{table}

\begin{figure*}[!t]
    \centering
    \includegraphics[width=\textwidth]{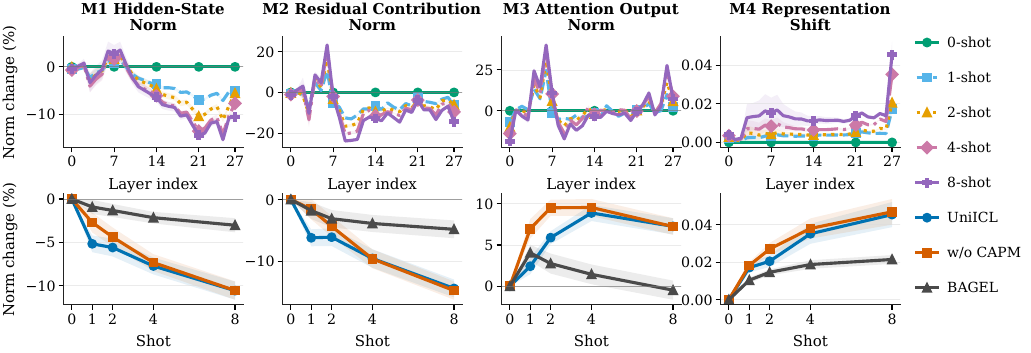}
    \caption{\textbf{Forward-feature control study on visual-grounding cases.} The upper row shows layerwise changes across shots, and the lower row tracks final-layer changes across models. Demonstrations mainly alter late-layer representations rather than uniformly amplifying activations.}
    \label{fig:feature_control}
\end{figure*}

\begin{figure}[!t]
    \centering
    \includegraphics[width=\columnwidth]{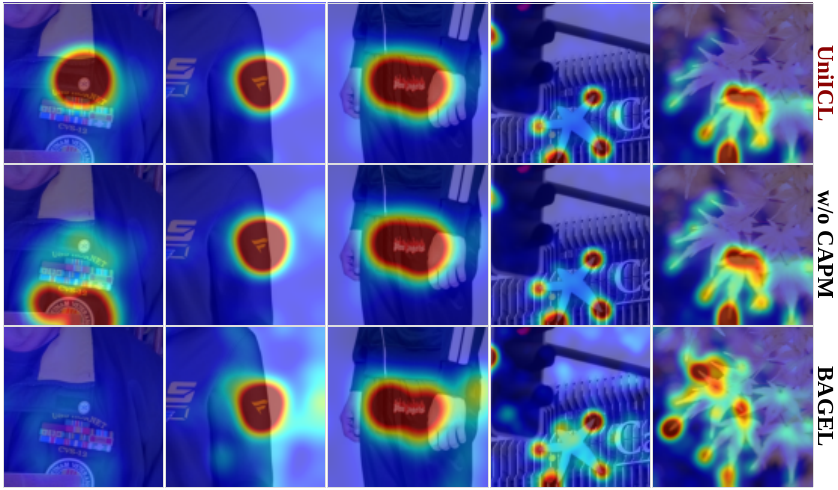}
    \caption{\textbf{Layer-20 query-to-image attention maps on five \(4\)-shot visual-grounding examples.} UniICL more consistently keeps high activation on the queried target, while the baselines diffuse attention to nearby salient regions.}
    \label{fig:feature_attention}
\end{figure}

\subsection{Feature-Level Analysis}
We examine the same effect inside the model in \cref{fig:feature_control,fig:feature_attention}. We run the forward-feature control on the same 20 visual-grounding cases under \(K\in\{0,1,2,4,8\}\) shots and compare \method{}, \method{} without CAPM, and BAGEL. For layer \(\ell\), we track four signals: hidden-state norm \(M_1\), residual contribution norm \(M_2\), attention-output norm \(M_3\), and representation shift \(M_4\). The first three are reported as relative changes from the matched zero-shot forward,
\(\Delta M_i^\ell(K)=100(M_i^\ell(K)-M_i^\ell(0))/M_i^\ell(0)\), while \(M_4^\ell(K)=1-\cos(h_\ell^K,h_\ell^0)\). The w/o-CAPM run is loaded as a separate model without constructing CAPM layers, so it is not a masked forward pass. For the attention comparison, we use \(4\)-shot episodes because CAPM is inactive at zero-shot.
\textbf{\textit{(1) Where demonstrations act.}} Shot scaling concentrates in the late layers. In the upper row of \cref{fig:feature_control}, the early and middle layers stay close across shots, while the final layers show monotonic growth in representation shift \(M_4\) as \(K\) increases. At the final layer, the hidden-state and residual norms decrease with more shots, while the attention-output branch is reweighted. The change is therefore directional, not a uniform activation increase. This locates the mechanism behind the stability results: demonstrations reshape the late representation instead of injecting a global signal. The shift magnitude only measures how strongly the model engages the demonstrations. Whether that engagement helps localization depends on how the shifted representation is routed.
\textbf{\textit{(2) Engagement versus routing.}} The lower row of \cref{fig:feature_control} tracks final-layer responses across shots. \method{} and w/o CAPM show similar representation-shift curves because they share the trained backbone and data, which explains why w/o CAPM remains a strong baseline. BAGEL differs in engagement: its final-layer shift stays about half of \method{} at \(4\)- and \(8\)-shot, and its attention-output change does not co-vary with the late representation. CAPM therefore does not simply enlarge the shift. It improves where the shift is routed. In \cref{fig:feature_attention}, \method{} keeps the high-activation region on the queried target, w/o CAPM often remains close but spreads more to salient neighboring regions, and BAGEL reaches the coarse area yet misses fine localization. The two failure modes are distinct: BAGEL under-engages the demonstrations, while w/o CAPM engages them with less reliable routing. CAPM mainly reduces the second error.

\begin{table*}[t]
    \centering
    \caption{\textbf{Human alignment of benchmark metrics.} Spearman's $\rho$ compares human preferences with automatic scores. A dash marks objective metrics that do not require pairwise human alignment.}
    \label{tab:human_study}
    \footnotesize
    \setlength{\tabcolsep}{3pt}
    \renewcommand{\arraystretch}{1.2}
    \begin{tabular}{@{}l|ccc|ccc|cc|cc|cc|ccc@{}}
        \toprule
        Level & \multicolumn{3}{c|}{Perception} & \multicolumn{3}{c|}{Imitation} & \multicolumn{2}{c|}{Conception} & \multicolumn{2}{c|}{Deduction} & \multicolumn{2}{c|}{Analogy} & \multicolumn{3}{c}{Discernment} \\
        Sub-task & VG & AR & IM & SAC & SR & IG & FCM & FCG & WAP & CoE & AI & AE & AA & FD & VR \\
        Metric & mIoU & Acc. & MLLM & MLLM & MLLM & HPSv3 & Acc. & MLLM & Acc. & MLLM & MLLM & MLLM & SRCC & Acc. & Q-Align \\
        \midrule
        $\rho$ & -- & -- & 0.812 & 0.828 & 0.867 & 0.913 & -- & 0.817 & -- & 0.835 & 0.826 & 0.809 & -- & -- & 0.887 \\
        \bottomrule
    \end{tabular}
\end{table*}

\begin{table}[t]
    \centering
    \caption{\textbf{Primary and auxiliary metric agreement.} Spearman's $\rho$ is computed within each sub-task.}
    \label{tab:metric_corr}
    \footnotesize
    \setlength{\tabcolsep}{4pt}
    \renewcommand{\arraystretch}{1.2}
    \begin{tabular*}{\columnwidth}{@{\extracolsep{\fill}}llcc@{}}
        \toprule
        Sub-task & Primary & Auxiliary & $\rho$ \\
        \midrule
        Style-Aware Caption  & MLLM-Judge & BERTScore & 0.612 \\
        Scene Reasoning      & MLLM-Judge & BERTScore & -0.016 \\
        Analogical Editing   & MLLM-Judge & DINOv3    & 0.746 \\
        Aesthetic Assessment & SRCC       & PLCC      & 0.914 \\
        \bottomrule
    \end{tabular*}
\end{table}

\subsection{Human Alignment}
To validate that our subjective metrics reflect human perception, we recruit $15$ evaluators to perform blind pairwise comparisons between \method{} and BAGEL across $350$ episodes from \benchmark spanning every capability level under the $2$-shot setting. Chain-of-Editing follows its native full-chain protocol. For each episode, evaluators assign a win score of $1$, a loss score of $0$, or a tie score of $0.5$ under three dimensions: Semantic Intent, Image Quality, and Aesthetics. Model identities are hidden and output order is shuffled per episode. For each subjective metric, we aggregate the pairwise labels into episode-level human preference scores for \method{} over BAGEL and correlate them with the corresponding automatic score margins using Spearman's $\rho$. Fleiss' $\kappa$ measures inter-annotator reliability. Objective metrics such as mIoU, Accuracy, and SRCC are excluded since they require no preference alignment. As shown in \cref{tab:human_study}, all subjective primary metrics align strongly with human preferences, with $\rho\in[0.81,0.91]$ and $\kappa{=}0.77$. Together with \cref{tab:metric_corr}, this justifies using task-specific scorers: BERTScore~\cite{zhang2019bertscore} is useful for form-sensitive captioning but unreliable for open-ended reasoning, while HPSv3, Q-Align, and MLLM-Judge show strong human alignment on their matched tasks.

\section{Conclusion}
\label{sec:conclusion}

We introduce {\method}, a paradigm for adapting multimodal understanding and generation models through demonstrations without per-task parameter updates. To address few-shot fragility, we propose a six-level Capability-Oriented Taxonomy and construct {\dataset} and {\benchmark}. The central finding of our analysis is that \emph{which} demonstrations are assembled matters far more than how many are given, and our curated, taxonomy-guided assembly pipeline is the primary driver of the observed gains. Our study of scaling behavior reveals non-monotonic effects: demonstrations can hinder perception via interference or enhance reasoning through inductive structure. As a complementary lightweight stabilizer, we additionally propose the Context-Adaptive Prototype Modulator, which provides further few-shot stability on top of these data-driven gains. On this data-centric foundation, our approach achieves competitive performance across unified baselines and remains competitive with specialized, larger-parameter models.

Additional task details, data-construction protocols, full results, implementation details, qualitative examples, and code-release files are provided in the Supplementary Material.

\noindent\textbf{Limitations and future work.} Dense visual generation tasks such as Image Manipulation remain vulnerable to context overload because overlapping edits require both rule transfer and precise pixel-level grounding. Our results reduce this degradation relative to baselines but do not eliminate it. The framework is currently limited to image-text modalities. Extending to video and audio would require handling temporal dynamics and cross-modal synchronization. The data pipeline depends on external foundation models, which may introduce biases. Finally, our few-shot regime covers $0$--$8$ shots, and long-context settings may reveal different scaling behaviors. Future work will extend the taxonomy and benchmark to broader modalities and study long-context few-shot stability.

\backmatter

\section*{Declarations}

\noindent\textbf{Competing interests.} The authors declare no competing interests.

\noindent\textbf{Data availability.} The dataset is available at \url{https://huggingface.co/datasets/xuyicheng-zju/UniICL-760K}.

\noindent\textbf{Code availability.} The code is available at \url{https://github.com/xuyicheng-zju/UniICL}.

\bibliography{references}

\end{document}